\documentclass[journal]{IEEEtran}
\usepackage{times}
\usepackage{epsfig}
\usepackage{graphicx}
\usepackage{amsmath}
\usepackage{amssymb}
\usepackage{float}
\usepackage{multirow}
\usepackage{bbding}
\usepackage{subfigure}

\usepackage{color}

\ifCLASSINFOpdf
\else
\fi
\begin{document}
%
\title{
Towards more realistic human motion prediction with attention to motion coordination}
%
%
%

\author{Pengxiang~Ding,~Jianqin~Yin, ~\IEEEmembership {Member, IEEE}
\thanks{
Pengxiang Ding and Jianqin Yin are with the School of
Artificial Intelligence of Beijing University of Posts and Telecommunications,
No.10 Xitucheng Road, Haidian District, Beijing 100876, China. E-mail:
dingpx2015@bupt.edu.cn, jqyin@bupt.edu.cn.

Corresponding author: Jianqin Yin.
}
}

%
%

\markboth{Copyright © 2022 IEEE. Personal use of this material is permitted. However, permission to use this material for any other purposes must be obtained from the IEEE by sending an email to pubs-permissions@ieee.org.}%
{Shell \MakeLowercase{\textit{et al.}}: bare Demo of IEEEtran.cls for IEEE Journals}
%



\maketitle

\begin{abstract}
Joint relation modeling is a curial component in human motion prediction. Most existing methods rely on skeletal-based graphs to build the joint relations, where local interactive relations between joint pairs are well learned. However, the motion coordination, a global joint relation reflecting the simultaneous cooperation of all joints, is usually weakened because it is learned from part to whole progressively and asynchronously. Thus, the final predicted motions usually appear unrealistic. To tackle this issue, we learn a medium, called coordination attractor (CA), from the spatiotemporal features of motion to characterize the global motion features, which is subsequently used to build new relative joint relations. Through the CA, all joints are related simultaneously, and thus the motion coordination of all joints can be better learned. Based on this, we further propose a novel joint relation modeling module, Comprehensive Joint Relation Extractor (CJRE), to combine this motion coordination with the local interactions between joint pairs in a unified manner. Additionally, we also present a Multi-timescale Dynamics Extractor (MTDE) to extract enriched dynamics from the raw position information for effective prediction. Extensive experiments show that the proposed framework outperforms state-of-the-art methods in both short- and long-term predictions on H3.6M, CMU-Mocap, and 3DPW.
\end{abstract}

\begin{IEEEkeywords}
human motion prediction, joint relation modeling, motion coordination, enriched dynamics
\end{IEEEkeywords}

%
\IEEEpeerreviewmaketitle

\section{Introduction}

Human motion prediction aims to generate future skeleton sequences from given past observed ones. It is a crucial research field since it can help machines respond rapidly to unknown situations in the future. Thus this technique has attracted much attention in many scenarios, such as human-robot interaction \cite{00}, \cite{-01}, \cite{03,-08}, autonomous driving \cite{01}, and pedestrian tracking \cite{02,05}. 

Predicting human motion is a challenging task because human motion is highly dynamic, non-linear, and more stochastic over time. Traditional works \cite{-02,-03} extend classical sequential models \cite{12,13} to motion prediction and achieved good performance in simple actions. Recently, those learning-based methods \cite{-04}, \cite{   16,17,18,21,22,19,20,23,33} have proven to be more effective in the complicated motions due to their better nonlinear fitting ability. They are mainly divided into two categories: RNN-based methods \cite{-04}, \cite{16,17,18,21,22} and Feed-forward methods\cite{19,20,23,08,33,34}. RNN-based methods usually suffer from some inherent problems existing in RNN, e.g., the prediction discontinuities and error-accumulation. Differently, feed-forward methods, including CNNs \cite{20}, \cite{19} and GCNs \cite{23,08,33,34}, can help alleviate those problems because the whole prediction process is not recursive. Besides, compared with RNN-based methods, feed-forward methods can model spatial and temporal relations simultaneously and exploit richer motion features. Thus, recent SOTA approaches mainly adopt feed-forward neural networks, and our work also falls under this category.

\begin{figure}[htb] 
\begin{center}
   \includegraphics[width=0.5\textwidth]{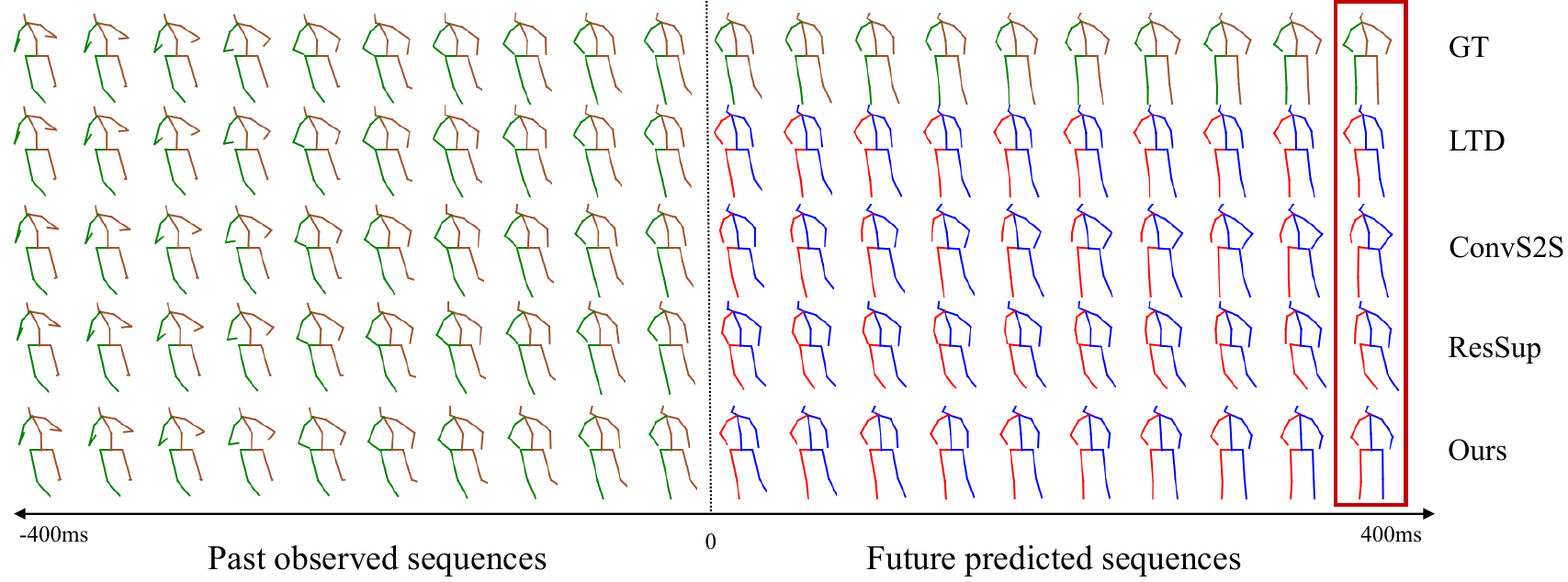} 
\end{center}
\caption{Qualitative results of short-term predictions of motion ``discussion'' on H3.6M. From top to bottom, we show the ground truth, the results of LTD \cite{08}, ConvS2S \cite{19}, ResSup \cite{17} and our approach. Compared with the result of our approach, the predicted motions of other works have the same problem: the limbs are uncoordinated which makes the predicted motion appear unrealistic.}
\label{f_vis_all}
\end{figure}

Although promising results are achieved in previous works, there is still a universal problem in previous works: they mainly focus on the local interactive relations between joint pairs but overlook the motion coordination, a specific global joint relation that encodes the simultaneous cooperation of all joints. In prior works, joint relations are usually modeled according to skeletal structure \cite{20}, \cite{23,22,21} or dynamic graphs \cite{34,08,33}. They can exploit rich local joint relations between both spatial-connected and spatial-separated joint pairs. However, global joint relations are usually learned by fusing different body components’ local motion features. In this way, the learned global joint relations can’t reflect the simultaneous cooperation of all joints, and thus the final predicted motion usually appears unrealistic, e.g., the arms and legs are uncoordinated, as is shown in Fig. \ref{f_vis_all}.

Therefore, in this paper, we propose to model motion coordination by exploring the simultaneous cooperation of all joints. To this end, we design a medium, called Coordination Attractor (CA), to correlate all joints simultaneously in an indirect way. In particular, the CA is learned by calculating the feature aggregation of all joints to represent the global motion features, which is next used to generate new relative joint features by subtracting raw features of each joint. In this way, all joints are correlated through the CA and generate new relative joint relations. Especially, the new relative joint relations enhance the cooperative relations of joints by reducing the motion commonality of joints to exploit more pure motion coordination. And then, the resulting relative joint features are next used to calculate joints similarities to generate final joint relations. In this way, all joints are correlated simultaneous in an indirect way, and thus the motion coordination can be modeled explicitly. Furthermore, the global motion coordination of all joints is not used alone but is combined with the local interactions between joint pairs to exploit richer joint relations for more accurate and realistic predictions.

\begin{figure*}[htb] 
	\begin{center}
		\includegraphics[width=1\textwidth]{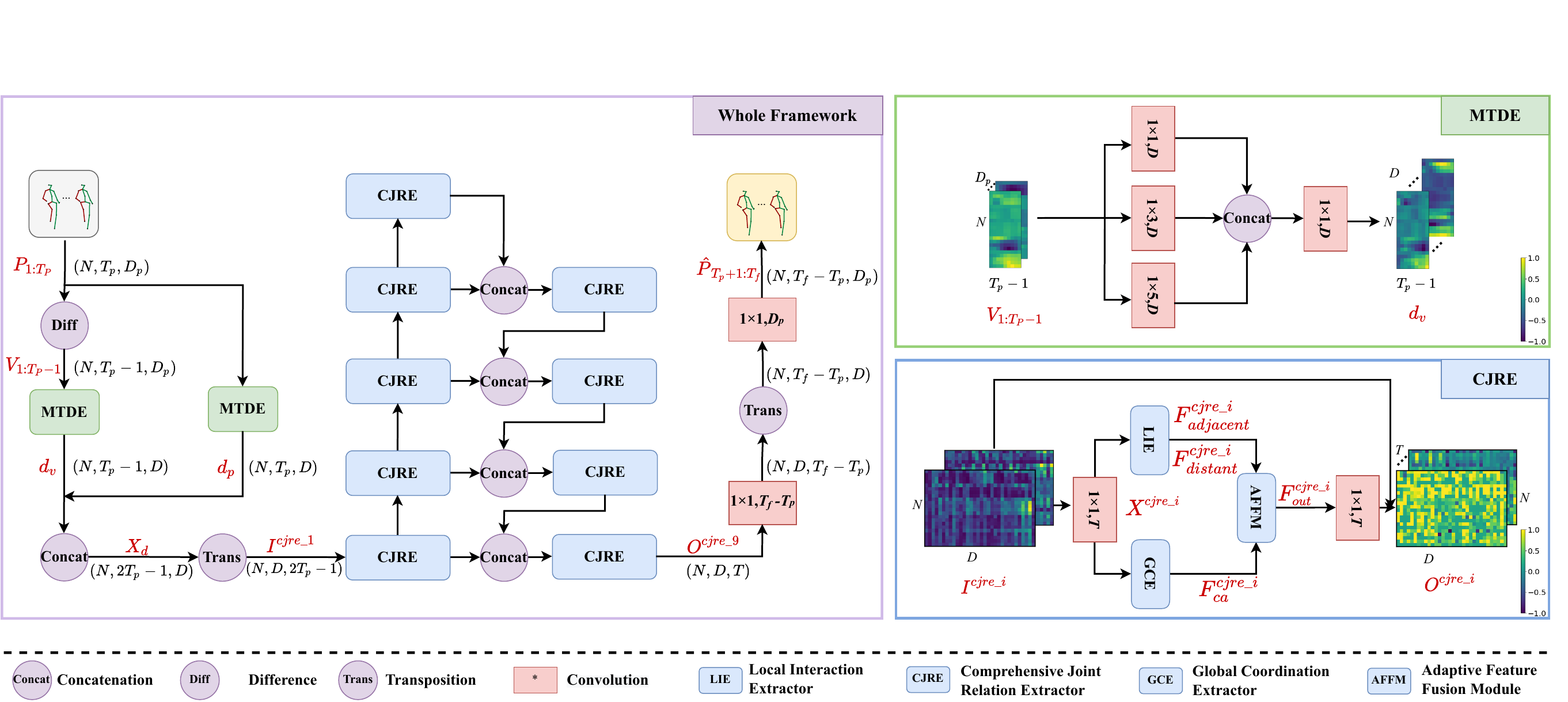} 
	\end{center}
	\caption{The left panel describes the whole framework of our proposed framework and the right two panels represent the details of MTDE and CJRE. Based on the two-stream architecture, the MTDE module is used to extract the enriched motion dynamics. The CJRE module is adapted to encode the global coordination of all joints and local interactions between joint pairs through GCE and LIE, respectively. We here denote ${I^{cjre\_i}}$ and ${O^{cjre\_i}}$ as input and output of the $i$th CJRE module. AFFM is introduced to fuse features according to the channel-wise attention mechanism. The whole CJRE is built based on the bottleneck architecture of ResNet \cite{27} for efficiency. Especially, lateral connections are used to offer fine-grained motion features inspired by U-Net \cite{28}. At last, two $1\times 1$ convolutions are successively used to transform temporal and spatial dimensions to get final prediction results.}
	\label{f_all}
\end{figure*}

Additionally, it is beneficial to exploit the enriched motion dynamics for effective prediction. As is well known, the skeleton sequences only include the position information of joints, which is insufficient to convey the dynamics of motion. Previous works \cite{10,31} tended to introduce extra velocity information to represent the motion dynamics via the two-stream architecture. However, the velocity only extracts the dynamics from neighbor frames, and thus more fine-grained dynamics is hard to capture simply from velocity. Therefore, in this paper, we propose to enrich the joint representation by extracting more diverse dynamics existing in different timescales.

Based on the above two aspects, we present our framework to generate more realistic and accurate human motions. Given observed motion sequences, we first learn enriched dynamics from raw position information adaptively through Multi-timescale Dynamics Extractor (MTDE). Next, we introduce the Comprehensive Joint Relation Extractor (CJRE), including a Local Interaction Extractor (LIE), a Global Coordination Extractor (GCE), and an Adaptive Feature Fusing Module (AFFM). The GCE is designed to present the global coordination of all joints, and the LIE is used to encode the local interactions between joint pairs. The above different joint relations are adaptively aggregated in the Adaptive Feature Fusing module (AFFM). Especially, we utilize the lateral connections between CJRE blocks for more fine-grained motion features. In this way, our proposed framework can capture more comprehensive joint relations and generate more diverse motion features for realistic and accurate predictions.

The main contributions of this paper are summarized as follows.

\begin{itemize}

\item We present to model motion coordination, a specific global joint relation that encodes the simultaneous cooperation of all joints, to enhance the realness of predicted motion. This motion coordination is further combined with the local interactions between joint pairs in a unified joint relation modeling module, CJRE, to extract richer joint relations for more realistic and accurate predictions.
\item We also put forward an MTDE module to extract enriched dynamics from the raw input data for effective prediction.
\item Our proposed framework outperforms most state-of-the-art methods for short and long-term motion prediction on three standard benchmark datasets: H3.6M, CMU-Mocap, and 3DPW.
\end{itemize}

\section{Related work}

Skeleton-based motion prediction has attracted increasing attention recently. Recent works using neural networks \cite{-04,-05,-06,-07,-09,08,16,17,18,19,20,21,22,23,34,33,32,38,35} have significantly outperformed traditional approaches \cite{-02,-03}.

\textbf{Human motion prediction.} RNNs \cite{16,17,18} are first used to predict human motion for their ability on sequence modeling. The first attempt was made by Fragkiadaki et al. \cite{16}, who proposed an Encoder-Recurrent-Decoder (ERD) model to combine encoder and decoder with recurrent layers. They encoded the skeleton in each frame to a feature vector and built temporal correlation recursively. Julieta et al. \cite{17} introduced a residual architecture to predict velocities and achieved better performance. Chen et al. \cite{-04} propose to modify the RNN structure using a novel diffusion convolutional recurrent predictor to model spatialtemporal motion features for better prediction. However, these works all suffer from discontinuities between the observed poses and the predicted future ones. Though Gui et al. \cite{18} proposed to generate a smooth and realistic sequence through adversarial training, it is hard to alleviate error-accumulation in a long-time horizon inherent to the RNNs scheme. A feedforward network was widely adopted to help alleviate those above questions because its prediction was not recursive and thus could avoid error-accumulation. Li et al. \cite{19} introduced a convolutional sequence-to-sequence model that encodes the skeleton sequence as a matrix whose columns represent the pose at every time step. However, their spatiotemporal modeling is still limited by the convolutional filters' size. Recently, \cite{20}, \cite{08} were proposed to consider global spatial and temporal features simultaneously. They all transformed temporal space to trajectory space to take the global temporal information into account. It contributes to capturing richer temporal correlation and thus achieved state-of-the-art results. In this paper, we follow this scheme but use different methods to model the spatial  relations of joints.

\textbf{Joint relation modeling.} 
Previous work mainly focused on skeletal constraints to model correlations between joints. Jain et al. \cite{21} first introduced a Structural-RNN model to explicitly model structural information relying on high-level spatiotemporal graphs. However, the graph is designed according to kinetic structure and is not flexible for different motions. Recently, some dynamic graph structures \cite{08,32,23,36} were developed to model more flexible joint relations. Mao et al. \cite{08} used an adaptive graph to model motion, but it is still unreliable because the graph is initialized randomly without structure prior. Cai et al. \cite{32} further combined kinematic structure with dynamic graph structure. Li et al. \cite{23} used stacked GCNs to build the interaction of different scales structure in each layer to model the correlation of both neighbor and distant joints. However, those above learned joint relations mainly refer to the local interactions between joint pairs without considering global motion coordination of all joints, which usually makes the final predicted motion appear unnatural or unrealistic. Therefore, in this paper, we aim to model more comprehensive joint relations, including both global motion coordination of all joints and local interactions between joint pairs.

\textbf{Motion dynamics of skeleton sequences.} 
The raw skeleton sequences are insufficient to convey the dynamics of motion because they only represent each joint's position information at each time step. Many attempts \cite{10,11,31} proposed to extract enriched dynamic representation from raw input data. They tended to rely on the two-stream architecture to introduce velocity information. A drawback of these methods is that they only extract the dynamics from neighbor frames. Though Li et al. \cite{10} enlarged the time horizon by convolution operation, it is still insufficient because different timescales encode different dynamics. Therefore, in this paper, we propose to extract the enriched through multi-timescale convolutions from raw motion sequences.

\section{Our Method}

The proposed framework aims to generate more realistic and accurate human motions for effective prediction. The overall architecture is shown in Fig. \ref{f_all}. It mainly includes two components, Multi-timescale Dynamics Extractor (MTDE) and Comprehensive Joint Relation Extractor (CJRE). The MTDE extracts multi-timescale temporal information to achieve richer motion dynamics for prediction. The CJRE mines comprehensive joint relations to model the spatiotemporal evolution of human motion. There exist several components in the CJRE. The global Coordination Extractor (GCE) and Local Interaction Extractor (LIE) are proposed to model the global coordination of all joints and local interactions between joint pairs separately. The Adaptive Feature Fusion module (AFFM) is introduced to fuse different joint relations according to the channel-wise attention mechanism. Especially, the lateral connections between CJRE blocks are designed to get more fine-grained motion features for more accurate predictions. Finally, two $1\times 1$ convolutions are successively used to transform temporal and spatial dimensions to get final prediction.

\subsection{Problem formulation}

We denote the historical 3D skeleton-based poses as ${P}_{1:T_p}=\left[p_1,\cdots,p_{T_p}\right]\in\mathbb{R}^{N\times T_p\times D_p}$ and future poses as ${P}_{T_{p}+1:T_f} = \left[p_{T_p+1},\cdots,p_{T_f}\right]\in\mathbb{R}^{N\times (T_f-T_p)\times D_p}$, where $p_t\in\mathbb{R}^{N\times D_p}$ represents the 3D pose at time $t$ with $N$ joints. The $D_p=3$ depicts the dimension of joint coordinates. Our goal is to generate predicted poses, ${\hat{P}}_{T_{p}+1:T_f}=M(P_{1:T_p})$ through the proposed framework $M(\cdot)$.

\subsection{Multi-timescale Dynamics Extractor (MTDE)}

\begin{figure}[htb] 
\begin{center}
   \includegraphics[width=0.5\textwidth]{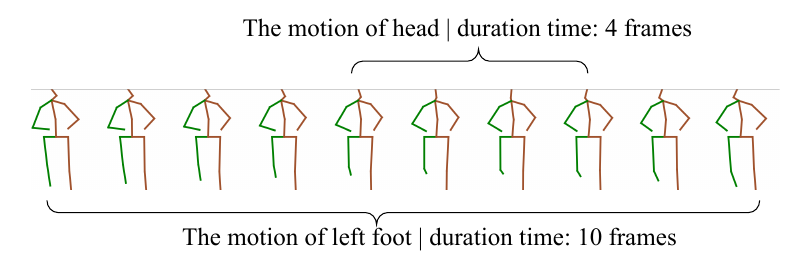} 
\end{center}
\caption{The obeservation of motion. It shows the duration time of the head is four frames, while the left foot is ten frames.}
\label{f_dynamics}
\end{figure}

Motion dynamics contains more motion cues compared with the position information in the raw motion sequences. It encodes the evolution of motion and thus is helpful to anticipate future motion trends. However, most previous works didn’t make use of this modality information. They tended to introduce the velocity as another input branch to enrich the input features and extract dynamics. Although it makes sense to some extent, it is insufficient only to use velocity to represent motion dynamics because more detailed and fine-grained dynamics couldn’t be captured simply from the velocity. We take Fig. \ref{f_dynamics} as an example. In this motion sequence, the duration time of the head movement is about four frames, while the left foot is ten frames. It shows two important details. First, the motion dynamics of different joints in a motion sequence usually are various. This observation reminds us that it is unsuitable to treat all joints equally, like the operation in calculating velocity. Second, unlike the velocity which only encodes the dynamics of adjacent frames, dynamics existing in different temporal scales could offer more diverse motion cues for accurate prediction.

\begin{figure*}[htb] 
\begin{center}
   \includegraphics[width=1\textwidth]{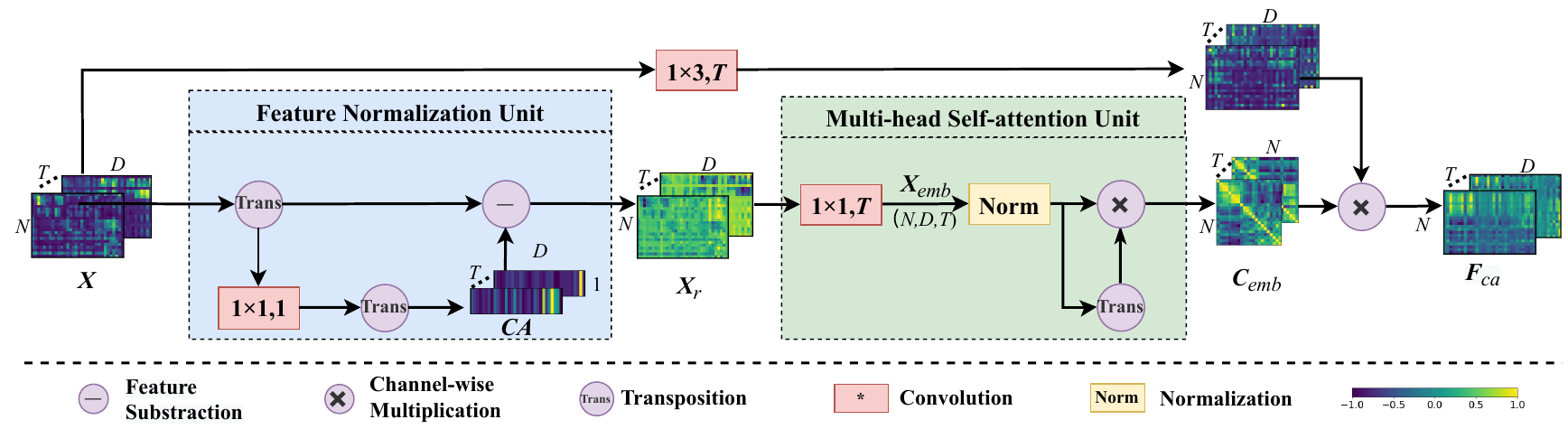} 
\end{center}
\caption{The overall architecture of GCE. It mainly contains two parts. The Feature Normalization Unit is designed to extract relative joint motion representation without the interference of global motion trends for later global coordination relation modeling. The Multi-head Self-attention Unit is proposed to generate multiple relation graphs of joints to extract richer global coordination. (To simplify the representation, ${X}$ and ${F_{ca}}$ are used to represent ${X^{cjre\_i}}$ and $F_{ca}^{cjre\_i}$, respectively.)}
\label{f_GCE2}
\end{figure*}

To address these issues, we propose two improvements based on the two-stream architecture in the Multi-timescale Dynamics Extractor (MTDE). Firstly, we extend the feature dimensions based on the raw joint position and velocity. In this way, we can extract more dynamic motion cues in high dimensional space, which bare position and velocity in 3D space can’t provide. Secondly, considering the dynamics of different joints are different even if in the same motion, we here apply multiple temporal convolutions to model more diverse motion dynamics existing in different temporal scales. Notably, we here only learn intra-joint dynamics to extract the motion dynamics without the interference of other joints. In this way, our proposed MTDE could extract richer motion dynamics to enrich intra-joint features, which is also beneficial for later inter-joint relation modeling. 

As is shown in Fig. \ref{f_all}, for the input $P_{1:T_p}\in\mathbb{R}^{N\times T_p\times D_p}$, we first get the velocity of raw motion sequence $V_{1:T_p-1}=[v_1,\cdots,v_{T_p-1}]\in\mathbb{R}^{N\times {(T_p-1)}\times D_p}$ by calculating the difference between adjacent frames of $P_{1:T_p}$. These two different branches $P_{1:T_p}$ and $V_{1:T_p-1}$ are all connected to a MTDE module to encode enriched dynamics through multi-timescale temporal convolutions respectively. Formally,

\begin{equation} \label{eqn_1}
  \begin{split}
  v_t&= (p_{t+1}-p_t),t=1,... ,T_p-1\\
  d_{k_i}^p&=\sigma({W}_{k_i}^p\ast P_{1:T_p}),i=1,2,3\\
  d_{k_i}^v&=\sigma({W}_{k_i}^t\ast V_{1:T_p-1}),i=1,2,3\\
  d_p&=W_p \ast Concat([d_{k_1}^p,d_{k_2}^p,d_{k_3}^p]), \\
  d_v&=W_v \ast Concat([d_{k_1}^v,d_{k_2}^v,d_{k_3}^v]),\\
  X_d&= d_p \oplus d_v\\
  \end{split}
\end{equation}

Here, the $\sigma(\cdot)$ is the activation function. ${W}_{k_i}^p$ and ${W}_{k_i}^v$ indicate the different $1\times k_i$ temporal convolution kernel with different timescale $k_i$ (the size of output channel is $D$). $Concat(\cdot)$ is the concatenation operation along the channel. $W_p$ and $W_v$ denote the $1\times 1$ convolution kernels used to fuse multi-timescale dynamics. $\ast$ and $\oplus$ denote the convolution operator and concatenation along temporal dimensions. For different input branches $P_{1:T_p}$ and $V_{1:T_p-1}$, $d_p\in\mathbb{R}^{N\times T_p\times D}$ and $d_v\in\mathbb{R}^{N\times (T_p-1) \times D}$ encode enriched dynamics existing in different timescales through fusing the features from different temporal convolutions respectively. To make use of different features of two branches, we synthesize $d_p$ and $d_v$ along temporal dimensions to get the representation $X_d\in\mathbb{R}^{N\times \left(2T_p-1\right)\times D}$. This operation enables the model to capture more detailed and fine-grained motion cues for later prediction.

\subsection{Global Coordination Extractor (GCE)} 

The global coordination of all joints plays an essential role in human motion. It describes the mutual constraints of all joints during motion and thus could offer richer motion cues to predict human motion. However, previous works mainly focused on modeling local interactions of joint pairs, and thus the global coordination wasn’t exploited well. Besides, there exist two major observations as to the global coordination of all joints. First, the global coordination essentially reflects the relative relations of all joints without global motion trends, which is not explored in previous works. Second, the learned global relations in most previous works are predefined and fixed, which is insufficient to represent the diversity of global coordination, such as balance, inertia, etc.

As to the above two observations, we propose the global Coordination Extractor (GCE) to model global coordination of all joints. It mainly contains two important components: Feature Normalization Unit and Multi-head Self-attention Unit. In the Feature Normalization Unit, we aim to extract relative joint motion representation without the interference of global motion trends for later global coordination relation modeling. In the Multi-head Self-attention Unit, multiple relation graphs of joints are built by calculating the self-attention of the learned relative joint representation. Notably, we learn the multiple relation graphs to combine different relative relations to extract richer joint relations. Next, we will illustrate more details in Fig. \ref{f_GCE2}. Notably, in the following part, all of the variables are simplified by removing superscript "${cjre\_i}$". For example, we use ${X}$ represents the $i$th CJRE module's variable ${X^{cjre\_i}}$.

\subsubsection{Definition of global motion trends}
As we discussed above, the global motion trends in this paper can be regarded as that the trajectory of global coordination center. Notably, we here learn it in high-dimensional trajectory space instead of in 3D space. The reason is that it is better to exploit more temporal consistency directly on trajectory space than 3D space, as illustrated in TrajCNN\cite{20}. To better introduce the trajectory of global coordination center, we first define it in the 3D space mathematically. Specifically, we denote the pose sequence as ${P}_{1:T_p}=\left[p_1,\cdots,p_{T_p}\right]\in\mathbb{R}^{N   \times T_p \times D_p}$ where $p_i$ represents the pose of $i$th frame. The global coordination center $cc_i$ of each pose can be regarded as the collective effect of all joints. Notably, considering that a skeleton is a non-rigid object, we use nonlinear transformation $nt(\cdot)$ to achieve $cc_i$, which is
formulated by formula \ref{eqn_2}.

\begin{equation} \label{eqn_2}
  \begin{split}
  {p}_{i} &= [j_1,...,j_N] \in\mathbb{R}^{N\times D_p}\\
  W &=[w_1,...,w_N] \in\mathbb{R}^{1\times N}\\
  {cc}_{i}&=nt({p}_{i})=\sigma(W \times {p}_{i}) \in\mathbb{R}^{1 \times D_p}\\
  \end{split}
\end{equation}

where $\sigma(\cdot)$ is the activation function and $W$ is the weight matrix. In this way, the trajectory of global coordination center of all frames can be represented as ${CC}=nt({P}_{1:T_p}^{'})=\left[cc_1,\cdots,cc_{T_p}\right]\in\mathbb{R}^{1 \times D_p\times T_p}$. (Notably, we here use ${P}_{1:T_p}^{'}\in\mathbb{R}^{N   \times D_p \times T_p}$ through transposing the dimension $D_p$ and $T_p$ of ${P}_{1:T_p}$ to keep the dimension consistency.)

Next, we extend the definition from 3D space to trajectory space. Concretely, we first get the features $X = f({P}_{1:T_p})\in\mathbb{R}^{N \times D \times T}$ through the feature extractor $f$. Here $D$ and $T$ represent the size of new features dimensions on trajectory space. Similarly, we can use the nonlinear transformation $nt(\cdot)$ to get the trajectory features of the global coordination center $CA = nt(f({P}_{1:T_p}))\in\mathbb{R}^{1 \times D^{} \times T^{}}$. Notably, we name the trajectory features of the global coordination center in trajectory space as Coordination Attractor $CA$ to distinguish it with the trajectory of the global coordination center $CC$ in 3D space.

Furthermore, we illustrate the equivalent form of nonlinear transformation $nt$. Formally,

\begin{equation} \label{eqn_3}
  \begin{split}
  CA & = nt(X) \\
  & = Trans(\sigma(W_{ca} \ast  Trans(X)) \\
  \end{split}
\end{equation}

The $\sigma(\cdot)$ is the activation function. $W_{ca}$ is a $1\times 1 $ convolution kernel (the size of output channel is 1) and $\ast$ denotes the convolution operator. We use $Trans(\cdot)$ to transpose the joint dimension and the temporal dimension. Because the size of the input channel is $N$ and the size of the output channel is 1, the output of the convolution operator is the global response of the $N$ joint features. In other words, the convolution operator computes the average weight of $N$ feature maps, which is the equivalent form of the nonlinear transformation $nt(\cdot)$ listed on formula  \ref{eqn_2}. We also use another $Trans(\cdot)$ to transpose the joint dimension and the temporal dimension back.

\subsubsection{Feature Normalization Unit}
This module aims to generate relative joint representation without global motion trends for later coordination modeling, as is illustrated in Fig \ref{f_GCE2}. Specifically, we first learn the Coordination Attractor ($CA$) to characterize the global motion features according to formula  \ref{eqn_3}. Subsequently, $CA$ is used as a medium to build a new relative joint representation $X_{r}\in\mathbb{R}^{N\times D\times T}$ indirectly through conducting feature subtraction. Formally,

\begin{equation} \label{eqn_4}
  \begin{split}
  X_{r}&= {X- CA}\\
  \end{split}
\end{equation}

In this way, the effect of global motion trends can be removed and the global coordination of all joints can be better modelled.

\subsubsection{Multi-head Self-attention Unit}
We aim to generate multiple joint relations which reflect global coordination by measuring the similarities of new relative features of each joint. Specifically, the whole process are shown as follows:

\begin{equation} \label{eqn_5}
  \begin{split}
  X_{emb}&=\sigma({\rm W}_{emb} \ast X_r) =\{x^{emb}_t\}_{t=1}^T \\
  C_{emb} &= \{c^{emb}_t\}_{t=1}^T \\
  c^{emb}_t&=SelfAtt(x^{emb}_t) \\
  \end{split}
\end{equation}

The $\sigma(\cdot)$ is the activation function. $SelfAtt(\cdot)$ is the function to calculate the self-attention of all joints in each feature map. $W_{emb}$ is a $1\times 1 $ convolution kernel (the size of output channel is $T$) and $\ast$ denotes the convolution operator. $x^{emb}_t\in\mathbb{R}^{N\times D}$ and $c^{emb}_t\in\mathbb{R}^{N\times N}$ are the feature map of $X_{emb}\in\mathbb{R}^{N\times D\times T}$ and $C_{emb}\in\mathbb{R}^{N\times N\times T}$ respectively. For the new relative joint features $X_{r}$, we first use ${W}_{emb}$ to learn a specific embedding for relation modeling. Next, we aim to calculate the relative joint relations $C_{emb}$. Notably, we calculate the relative relations on each feature map because each feature map encodes specific spatiotemporal features and should focus on different joint relations. We take $x^{emb}_t$ as an example. For $x^{emb}_t$, each row vector represents the features of one joint. Therefore, we can calculate the cosine similarity between all row vector pairs to illustrate the correlation between joint pairs. The reasons why we choose cosine similarity are: (1) this metric contains angle information that corresponds to the physical relations of joints; (2) the value is limited into $[-1,1]$, which avoids the violent variance. In this way, we could generate diverse global coordination through multiple relations graphs.

The last step is to calculate the coordinated motion features according to the joint relations $C_{emb}$. Considering each feature map of $C_{emb}$ represents one specific 
coordination, we apply channel-wise multiplication to make use of different relation graphs. Specifically, $1\times 3$ convolution $W_{intra}$ is used to extract intra-joint features and then combine with the guidance of $C^{emb}$ to get the final features $F_{ca}\in\mathbb{R}^{N\times D\times T}$.

\begin{equation}
F_{ca}=C_{emb}\odot(\sigma(W_{intra}\ast X))
\label{eq_6}
\end{equation}
where $\odot$ represents channel-wise multiplication.

\subsection{Local Interaction Extractor (LIE)}

\begin{figure}[htp] 

\begin{center}
\includegraphics[width=0.3\textwidth]{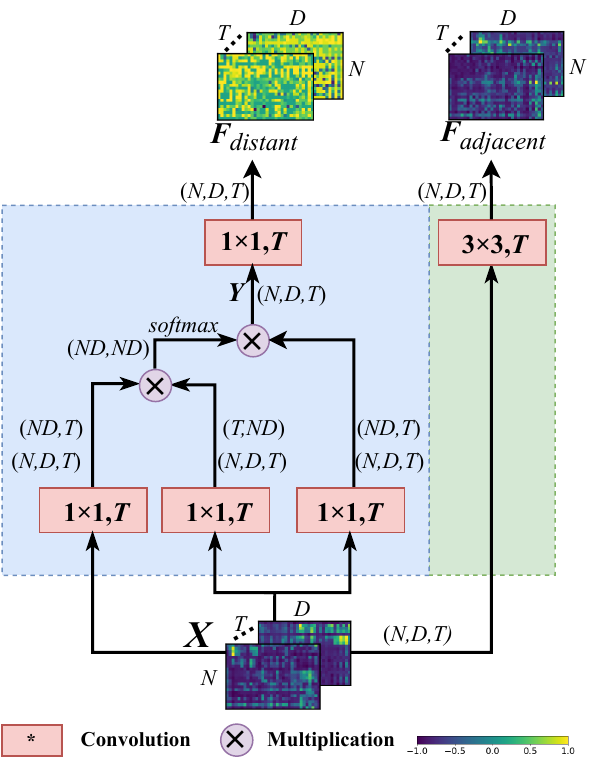} 
\end{center}
\caption{The implementation of Local Interaction Extractor (LIE). The left is the path using a Non-local block without residual connection to learn the relations between distant joint pairs. The right is the the path with convolutions to learn the relations between adjacent joint pairs. (To simplify the representation, ${X}$, ${F_{adjacent}}$ and ${F_{distant}}$ are used to represent ${X^{cjre\_i}}$, $F_{adjacent}^{cjre\_i}$, $F_{distant}^{cjre\_i}$ respectively.)
\label{f_LIE}}
\end{figure}

Local Interaction Extractor (LIE) is used to learn local interactions between joint pairs, including adjacent and distant joints. The local connection via bones brings spatial correlation for adjacent joints. For distant joints, some joints may have a strong correlation even if they are not directly connected, e.g., left hand and right hand are tightly correlated during ``eating''. Therefore, these two relations are equally important for effective prediction.

As is shown in Fig. \ref{f_LIE}, given an input $X\in\mathbb{R}^{N\times D\times T}$ which is the same as GCE, there exist two main paths to separately learn the relations between adjacent joint pairs and distant joint pairs. 

For the adjacent joints, we here use the $3\times 3$ convolution to combine the co-occurrence features of adjacent joints. In this way, local interactions between adjacent joints can be built.
\begin{equation}\label{eq_7}
	\begin{split}
		F_{adjacent}&=\sigma(W_{adjacent}\ast X)\\
	\end{split}
\end{equation}	

For the distant joints, considering the spatially separated joints are often far away from each other, we here adopt the self-attention mechanism used in the Non-local \cite{30} block to exploit their relations without the limitation of their arrangements in the feature maps.

Specifically, the input $X=[q_1,...,q_{N\times D}]\in\mathbb{R}^{N\times D\times T}$, where $q_i\in\mathbb{R}^{T}$ represents the features of ${i_{th}}$ pixel in all feature maps. 
To encode the relations between $i$ and different position $j$, we calculate the similarity of the $q_i$ and $q_j$ as the response of position $j$ to position $i$. In this way, the features of distant joints are correlated and the sum of all other position $j$’s response form the resulting response $y_i$. Formally,

\begin{equation}
\label{eq_14}
	\begin{split}
	g(q_i) &=\ W_g q_i \\
	f(q_i,q_j) &=softmax({\theta(q_i)}^T\varphi(q_j))\\
	y_i  &=\frac{1}{C(X)}\sum_{\forall j}{f(q_i,q_j)}g(q_j)\\
	\end{split}
\end{equation}

where $\theta$, $\varphi$, $g$ are $1\times 1$ convolutions and $i,j \in [1,...,N\times D]$. $C(X)$ is the normalization parameter. $\theta$ and $\varphi$ are used to learn the similarity between different positions. $g$ is used to learn an embedding representation of raw input X. The result $y_i$ is calculated by the  weighted sum of other positions' effect and finally constitutes the output $Y=[y_1,...,y_{N\times D}]\in\mathbb{R}^{N\times D\times T}$. In this way, the new feature $Y$ gets the relations of spatially separated joints without limiting their arrangements in the feature maps. And thus it can be used to calculate the final features of distant joints through the convolution kernel ${W_{distant}}$ for further feature extraction.

\begin{equation}
	\label{eq_15}
	\begin{split}
	F_{distant}&=\sigma(W_{distant}\ast Y)\\
	\end{split}
\end{equation}

\begin{table*}[!t] 
\caption{Short-term prediction of 8 sub-sequences per actionon on H$3.6$M. Where ``ms'' denotes ``milliseconds''.}
\scriptsize
\begin{center}
\begin{tabular}{c|cccc|cccc|cccc|cccc}
\hline
motion & \multicolumn{4}{c}{Walking} & \multicolumn{4}{c}{Eating}& \multicolumn{4}{c}{Smoking} & \multicolumn{4}{c}{Discussion}\\
\hline
time(ms)&80&160&320&400&80&160&320&400&80&160&320&400&80&160&320&400 \\
\hline
ResSup \cite{17} &23.8 &40.4& 62.9& 70.9& 17.6& 34.7& 71.9& 87.7& 19.7& 36.6& 61.8& 73.9&31.7& 61.3& 96.0& 103.5 \\
ConvS2S \cite{19} &17.1 &31.2&53.8&61.5&13.7&25.9&52.5&63.3&11.1&21.0&33.4&38.3&18.9&39.3&67.7&75.7\\
LTD \cite{08}&8.9 &15.7&29.2& 33.4& 8.8& 18.9& 39.4& 47.2& 7.8& 14.9& 25.3&{28.7}& 9.8& 22.1&{39.6} &{\bf44.1} \\

LPJP \cite{32} &7.9 &14.5&29.1&34.5&8.4&18.1&37.4&45.3&6.8&{13.2}&{24.1}&{\bf27.5}&8.3&{21.7}&43.9&48.0\\

TrajCNN \cite{20} &8.2 &14.9&30.0&35.4&8.5&18.4&37.0&44.8&6.3&{\bf12.8}&{\bf23.7}&{27.8}&7.5&{\bf20.0}&41.3&47.8\\

\hline
 Ours&{\bf 7.2} & {\bf 13.7} &{\bf25.6}& {\bf31.0}& {\bf 7.7} &{\bf 16.7}&{\bf 35.8}&{\bf 44.2} &{\bf 6.3}&{ 13.3} &{24.5}&29.7&{\bf 7.5} &{20.3}&{\bf 38.7}&{44.7}\\

\hline
\hline
motion & \multicolumn{4}{c}{Direction} & \multicolumn{4}{c}{Greeting}& \multicolumn{4}{c}{Phoning} & \multicolumn{4}{c}{Posing}\\
\hline
time(ms)&80&160&320&400&80&160&320&400&80&160&320&400&80&160&320&400 \\
\hline
ResSup \cite{17} & 36.5 &56.4& 81.5& 97.3&37.9& 74.1& 139.0& 158.8 &25.6& 44.4& 74.0& 84.2& 27.9& 54.7& 131.3& 160.8 \\
ConvS2S \cite{19} & 22.0&37.2 &59.6& 73.4 &24.5 &46.2 &90.0& 103.1& 17.2& 29.7& 53.4 &61.3& 16.1& 35.6& 86.2& 105.6\\

LTD \cite{08}& 12.6 & 24.4&{48.2}&{ 58.4}& 14.5& 30.5& 74.2& 89.0 & 11.5& 20.2& 37.9& 43.2&9.4& 23.9& {66.2}&{ 82.9}\\

LPJP \cite{32} &11.1 &22.7&48.0&58.4&13.2&28.0&64.5&77.9&10.8&{19.6}&{37.6}&{46.8}&8.3&{22.8}&65.6&81.8\\

TrajCNN \cite{20} &9.7 &22.3&50.2&61.7&12.6&28.1&67.3&80.1&10.7&18.8&37.0&43.1&6.9&21.3&62.9&78.8\\

\hline
Ours&{\bf 9.3}&{\bf 21.1}&{\bf 45.0}&{\bf 55.0}&{\bf 11.2}&{\bf 23.9}&{\bf 63.4}&{\bf 79.6}& {\bf 10.2}&{\bf 18.5}& {\bf 34.3}&{\bf 38.5}  &{\bf 6.8}&{\bf 20.5}&{\bf 60.6}&{\bf 76.6}\\

\hline
\hline
motion & \multicolumn{4}{c}{Purchasing} & \multicolumn{4}{c}{Sitting}& \multicolumn{4}{c}{Sitting down} & \multicolumn{4}{c}{Taking photo}\\
\hline
time(ms)&80&160&320&400&80&160&320&400&80&160&320&400&80&160&320&400 \\
\hline
ResSup \cite{17} & 40.8& 71.8& 104.2& 109.8 &34.5& 69.9& 126.3 &141.6& 28.6& 55.3& 101.6& 118.9 &23.6 &47.4& 94.0& 112.7\\
ConvS2S \cite{19} & 29.4& 54.9& 82.2& 93.0 &19.8 &42.4& 77.0& 88.4& 17.1& 34.9& 66.3& 77.7& 14.0& 27.2& 53.8& 66.2\\
LTD \cite{08}& 19.6&38.5&{ 64.4}&{ 72.2}&10.7& 24.6& 50.6&62.0&11.4 &{ 27.6}& 56.4& 67.6& 6.8& 15.2& {38.2}&{49.6} \\

LPJP \cite{32} &18.5 &38.1&{\bf61.8}&{\bf69.6}&9.5&23.9&49.8&61.8&11.2&{29.9}&{59.8}&{68.4}&6.3&{14.5}&38.8&49.4\\

TrajCNN \cite{20}& 17.1 &{\bf 36.1}&{64.3}&{75.1}& 9.0& 22.0& 49.4& 62.6 & 10.7& 28.8& 55.1& 62.9&{\bf5.4}& {\bf13.4}& {\bf36.2}&{\bf47.0}\\

\hline

Ours&{\bf 17.1}&38.0&65.0&73.0& {\bf 7.8}&{\bf 19.9}&{\bf 44.9}&{\bf 56.4}& {\bf 9.2}&{\bf 23.7}&{\bf 47.7}&{\bf 59.4}&{5.6}&{14.3}&{37.6}&{48.9}\\
\hline
\hline
motion & \multicolumn{4}{c}{Waiting} & \multicolumn{4}{c}{Walking dog}& \multicolumn{4}{c}{Walking Together} & \multicolumn{4}{c}{Average}\\
\cline{1-17}
time(ms)&80&160&320&400&80&160&320&400&80&160&320&400&80&160&320&400 \\
\hline
ResSup \cite{17} & 29.5& 60.5& 119.9& 140.6& 60.5& 101.9& 160.8& 188.3& 23.5& 45.0& 71.3& 82.8& 30.8& 57.0& 99.8& 115.5\\
ConvS2S \cite{19} &17.9& 36.5& 74.9& 90.7& 40.6& 74.7& 116.6& 138.7& 15.0& 29.9& 54.3& 65.8& 19.6& 37.8& 68.1& 80.2\\
LTD \cite{08}& 9.5& 22.0& 57.5& 73.9& 32.2& 58.0& 102.2& 122.7 & 8.9& { 18.4}& 35.3& 44.3& 12.1& 25.0& 51.0& 61.3\\

LPJP \cite{32} &8.4 &21.5&53.9&69.8&22.9&50.4&100.8&119.8&8.7&{18.3}&{34.2}&{44.1}&10.7&{23.8}&50.0&60.2\\

TrajCNN \cite{20}& 8.2 & 21.0&{53.4}&{68.9}& 23.6& 52.0& 98.1& 116.9 & 8.5& 18.5& 33.9& 43.4&10.2& 23.2& {49.3}&{ 59.7}\\

\hline
Ours&{\bf 7.7}&{\bf 18.8}&{\bf 48.0}&{\bf 64.7} & {\bf 22.0}&{\bf 49.2}&{\bf 90.9}&{\bf 110.0}&  {\bf 7.8}&{\bf 17.3}&{\bf 32.1}&{\bf 43.3}&{\bf 9.6}&{\bf 22.0}&{\bf 46.2}&{\bf 57.0}\\

\hline

\end{tabular}

\end{center}
\label{r_h36mshort}
\vspace{-2.5em}
\end{table*}

\subsection{Adaptive Feature Fusing Module (AFFM)}

The different motions will have a respective preference for local interactions between joint pairs and global coordination of all joints. For example, dynamic actions like 'walking' may pay more attention to the local interactions between joint pairs of which the movement are more obvious while static actions like 'sitting' may focus on the entire structure of all joints. Thus, we aim to measure the importance of different features in AFFM module.

\begin{figure}[htp] 
\begin{center}
\includegraphics[width=0.5\textwidth]{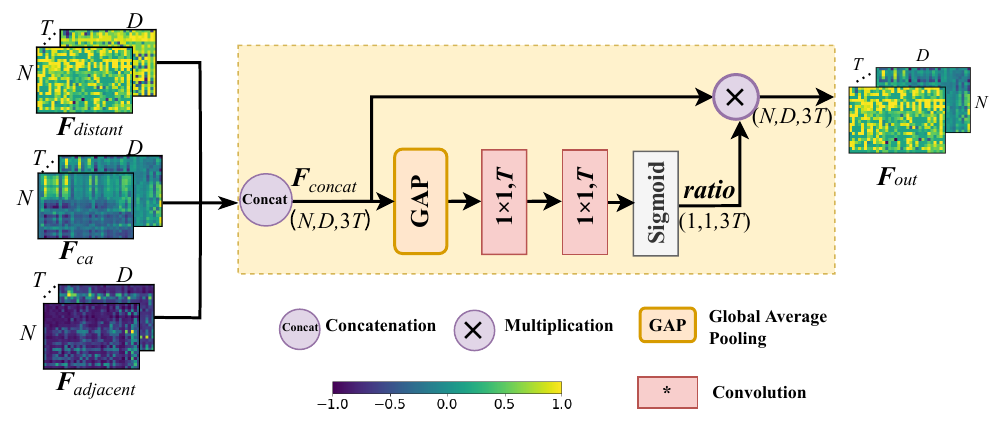} 
\end{center}

\caption{The implementations of Adaptive Feature Fusing Module (AFFM). The features learnt from previous block are fused with channel attention machanism. (To simplify the representation, ${F_{adjacent}}$, ${F_{distant}}$, ${F_{ca}}$, ${F_{out}}$ are used to represent $F_{adjacent}^{cjre\_i}$, $F_{distant}^{cjre\_i}$, $F_{ca}^{cjre\_i}$ and $F_{out}^{cjre\_i}$ respectively.)
\label{f_AFFM}}

\end{figure}

Specifically, as is shown in Fig. \ref{f_AFFM}, we first combine different features extracted from GCE and LIE along the channel dimension. The next global average pooling squeeze each feature map as a scalar. In this way, each feature map can be encoded into a global feature. Then to learn the importance ratio between channels, we use the nonlinear transformation by two $1\times1$ convolution layers and use a sigmoid layer to get the final importance ratio. Formally, 

\begin{equation}
    \label{eq_16}
	\begin{split}
	    F_{concat}&= Concat([F_{distant},F_{ca},F_{adjacent}])\\
	    ratio&=Sigmoid(W_{a2}\ast \sigma(W_{a1}\ast GAP(F_{concat})))\\
	    F_{out}&= ratio\odot F_{concat}\\
	\end{split}
\end{equation}	

where $Concat(\cdot)$ is the concatenation operation along the channel. $GAP$ is the global average pooling. $W_{a1}$ and $W_{a2}$ indicate the different convolution kernel to be used to nolinear transformation. The $\sigma(\cdot)$ is the activation function. The $Sigmoid(\cdot)$ function is used to limit the value of $ratio$ in $[0,1]$. The $\odot$ represents channel-wise multiplication. At last, channel attention mechanism can be used to conduct channel-wise multiplication between ratio and concatenated features $F_{concat}$ to reform new features. In this way, those features with higher important ratios can be enhanced and the later prediction could pay more attention to those enhanced features, which is beneficial to more precise prediction.

\begin{table*}[!t] 
\vspace{0.5em}
\caption{Long-term prediction of 8 sub-sequences per actionon on H$3.6$M. Where ``ms'' denotes ``milliseconds''.}
\scriptsize
\begin{center}
\begin{tabular}{c|cc|cc|cc|cc|cc|cc|cc|cc}
\hline
motion & \multicolumn{2}{c}{Walking} & \multicolumn{2}{c}{Eating}& \multicolumn{2}{c}{Smoking} & \multicolumn{2}{c}{Discussion}
&\multicolumn{2}{c}{Directions} & \multicolumn{2}{c}{Greeting}& \multicolumn{2}{c}{Phoning} & \multicolumn{2}{c}{Posing}\\
\hline
time(ms)&560 &1000&560 &1000&560 &1000&560 &1000&560 &1000&560 &1000&560 &1000&560 &1000\\
\hline
LTD \cite{08}&42.2&51.3&{\bf56.5}&{\bf68.6}&32.3&60.5&{\bf70.4}&103.5&85.8&109.3&91.8&87.4&65.0&113.6&113.4&220.6\\
TrajCNN \cite{20}&37.9&46.4&59.2&71.5&32.7&58.7&75.4&{\bf103.0}&{\bf84.7}&104.2&91.4&{\bf84.3}&62.3&113.5&111.6&{\bf210.9}\\
\hline
Ours &{\bf35.5}&{\bf42.7}  &57.3&70.3 &{\bf30.9}&{\bf55.0} &74.3&105.7 &89.7&{\bf103.5} &{\bf91.1}&90.5  &{\bf59.1}&{\bf110.5} &{\bf107.3}&211.9 \\
\hline
\end{tabular}
\begin{tabular}{c|cc|cc|cc|cc|cc|cc|cc|cc}
\hline
motion & \multicolumn{2}{c}{Purchases} & \multicolumn{2}{c}{Sitting}& \multicolumn{2}{c}{Sitting down} & \multicolumn{2}{c}{Taking photo}
&\multicolumn{2}{c}{Waiting} & \multicolumn{2}{c}{Walking Dog}& \multicolumn{2}{c}{Walking Together} &\multicolumn{2}{c}{Average}\\
\cline{1-17}
time(ms)&560 &1000&560 &1000&560 &1000&560 &1000&560 &1000&560 &1000&560 &1000&560 &1000\\
\hline
LTD \cite{08}&94.3&130.4&79.6&114.9&82.6&140.1&68.9&87.1&100.9&167.6&136.6 &174.3&{\bf57.0}&85.0&78.5&114.3\\
TrajCNN \cite{20}&84.5&{\bf115.5}&81.0&116.3&79.8&{\bf123.8}&{\bf73.0}&{\bf86.6}&92.9&165.9&141.1&181.3&57.6&{\bf77.3}&77.7&110.6\\
\hline
Ours&{\bf82.1}&117.6&{\bf73.1}&{\bf105.1}&{\bf78.0}&126.1&75.9&88.9&{\bf85.9}&{\bf154.4}&{\bf130.2}&{\bf170.7}&57.1&82.2&{\bf75.1}&{\bf109.0}\\
\hline
\end{tabular}
\end{center}
\label{r_h36mlong}
\vspace{-2.0em}
\end{table*}

\subsection{Loss Function}
Following \cite{20,08}, we make use of the Mean Per Joint Position Error (MPJPE). In particular, for one training sample, the loss is as follows:

\begin{equation}
L=\ \frac{1}{N\times\left(T_f-T_p\right)}\sum_{i=T_p+1}^{T_f}\sum_{j=1}^{N}{\parallel P_{i,j}-\ {\hat{P}}_{i,j}\parallel_2}
\label{eq_17}
\end{equation}
where ${\hat{P}}_{i,j}\in R^3$, representing the 3D coordinates of the $j$th joint of the $i$th human pose, is the predicted result and $P_{i,j}\in R^3$ is the ground truth.

\section{Experiments}
In this section, we first introduce the datasets used in our experiments and the implementation details of our work. Then, we compare our method with baselines. Next, we carry out some experiments to analyze the contributions of the proposed method and visualize the predictive performance of our model.

\subsection{Datasets and Implementation Details}

\begin{table*}[htb] 
\caption{Short-term prediction of 256 sub-sequences per actionon on H$3.6$M. Where ``ms'' denotes ``milliseconds''.}
\scriptsize
\begin{center}
\begin{tabular}{c|cccc|cccc|cccc|cccc}
\hline
motion & \multicolumn{4}{c}{Walking} & \multicolumn{4}{c}{Eating}& \multicolumn{4}{c}{Smoking} & \multicolumn{4}{c}{Discussion}\\
\hline
time(ms)&80&160&320&400&80&160&320&400&80&160&320&400&80&160&320&400 \\
\hline
ResSup \cite{17} &23.2 &40.9& 61.0& 66.1& 16.8& 31.5& 53.5& 61.7& 18.9& 34.7& 57.5& 65.4&25.7& 47.8& 80.0& 91.7 \\
ConvS2S \cite{19} &17.1 &33.5&56.3&63.6&11.0&22.4&40.7&48.4&11.6&22.8&41.3&48.9&17.1&34.5&64.8&77.6\\

LTD \cite{08}&11.1 &21.4&37.3& 42.9& 7.0& 14.8& 29.8& 37.3& 7.5& 15.5& 30.7&37.5& 10.8& 24.0&52.7 &65.8 \\
HRI     \cite{31}&10.0 &19.5&{\bf34.2}&{\bf 39.8}& 6.4& 14.0& 28.7& 36.2& 7.0& 14.9& 29.9&36.4& 10.2& 23.4&52.1 &65.4 \\

\hline{}
 Ours&{\bf 9.4} & {\bf 18.9} &34.5& 41.3& {\bf 5.6} &{\bf 13.0}&{\bf 27.5}&{\bf 35.2} &{\bf 6.2}&{\bf 13.7} &{\bf28.3}&{\bf35.4}&{\bf 8.8} &{\bf21.8}&{\bf 50.5}&{\bf63.8}\\
\hline
\hline
motion & \multicolumn{4}{c}{Direction} & \multicolumn{4}{c}{Greeting}& \multicolumn{4}{c}{Phoning} & \multicolumn{4}{c}{Posing}\\
\hline
time(ms)&80&160&320&400&80&160&320&400&80&160&320&400&80&160&320&400 \\
\hline
ResSup    \cite{17} &21.6 &41.3 &72.1 &84.1         & 31.2 &58.4 &96.3 &108.8           & 21.1 & 38.9 & 66.0 &76.4          & 29.3 & 56.1 &98.3 &114.3 \\
ConvS2S   \cite{19} &13.5 &29.0 &57.6 &69.7         & 22.0 &45.0 &82.0 &96.0            & 13.5 & 26.6 & 49.9 &59.9            & 16.9 & 36.7 &75.7 &92.9\\

LTD \cite{08} &8.0  &18.8 &43.7 &54.9         & 14.8 &31.4 &65.3 &79.7          & 9.3  & 19.1 & 39.8 &49.7          & 10.9 & 25.1 &59.1 &75.9 \\
HRI     \cite{31} &7.4  &18.4 &44.5 &56.5         & 13.7 &30.1 &63.8 &78.1            & 8.5  & 18.3 & 39.0 &49.2                  & 10.2 & 24.2 &58.5 &75.8 \\

\hline
 Ours&{\bf6.3}&{\bf16.9}&{\bf42.1}&{\bf54.0}    &{\bf11.9}&{\bf27.9}&{\bf61.3}&{\bf76.1}    &{\bf7.6}&{\bf17.1} &{\bf37.4}&{\bf47.6}    &{\bf8.4} &{\bf21.9}&{\bf54.8}&{\bf71.7}\\

\hline
\hline
motion & \multicolumn{4}{c}{Purchasing} & \multicolumn{4}{c}{Sitting}& \multicolumn{4}{c}{Sitting down} & \multicolumn{4}{c}{Taking photo}\\
\hline
time(ms)&80&160&320&400&80&160&320&400&80&160&320&400&80&160&320&400 \\
\hline
ResSup    \cite{17} &28.7 &52.4 &86.9 &100.7              & 23.8 &44.7 &78.0 &91.2            & 31.7 & 58.3 & 96.7 &112.0         & 21.9 & 41.4 &74.0 &87.6 \\
ConvS2S   \cite{19} &20.3 &41.8 &76.5 &89.9               & 13.5 &27.0 &52.0 &63.1            & 20.7 & 40.6 & 70.4 &82.7            & 12.7 & 26.0 &52.1 &63.6\\

LTD \cite{08} &13.9 &30.3 &62.2 &75.9               & 9.8  &20.5 &44.2 &55.9          & 15.6 & 31.4 & 59.1 &71.7            & 8.9  & 18.9 &41.0 &51.7 \\
HRI     \cite{31} &13.0 &29.2 &{\bf60.4} &{\bf73.9}           & 9.3  &20.1 &44.3 &56.0            & 14.9 & 30.7 & 59.1 &72.0                  & 8.3  & 18.4 &40.7 &51.5 \\

\hline
 Ours&{\bf11.3}&{\bf27.6}&60.6&74.9               &{\bf8.0}&{\bf18.1}&{\bf41.1}&{\bf52.7}    &{\bf13.0}&{\bf28.5} &{\bf56.5}&{\bf70.2}    &{\bf7.4} &{\bf17.0}&{\bf39.3}&{\bf50.2}\\
\hline
\hline
motion & \multicolumn{4}{c}{Waiting} & \multicolumn{4}{c}{Walking dog}& \multicolumn{4}{c}{Walking Together} & \multicolumn{4}{c}{Average}\\
\cline{1-17}
time(ms)&80&160&320&400&80&160&320&400&80&160&320&400&80&160&320&400 \\
\hline
ResSup    \cite{17} &23.8 &44.2 &75.8 &87.7         & 36.4 &64.8 &99.1 &110.6           & 20.4 & 37.1 & 59.4 &67.3          & 25.0 & 46.2 &77.0 &88.3 \\
ConvS2S   \cite{19} &14.6 &29.7 &58.1 &69.7         & 27.7 &53.6 &90.7 &103.3           & 15.3 & 30.4 & 53.1 &61.2            & 16.6 & 33.3 &61.4 &72.7\\

LTD \cite{08} &9.2  &19.5 &43.3 &54.4         & 20.9 &40.7 &73.6 &86.6          & 9.6  & 19.4 & 36.5 &44.0          & 11.2 & 23.4 &47.9 &58.9 \\
HRI     \cite{31} &8.7  &19.2 &43.4 &54.9         & 20.1 &40.3 &73.3 &86.3            & 8.9  & 18.4 & 35.1 &{\bf41.9}                 & 10.4 & 22.6 &47.1 &58.3 \\

\hline
 Ours&{\bf7.4}&{\bf17.0}&{\bf41.0}&{\bf52.3}    &{\bf17.4}&{\bf37.0}&{\bf71.2}&{\bf85.0}    &{\bf8.0}&{\bf17.1} &{\bf33.7}&{\bf41.9}    &{\bf9.1} &{\bf21.0}&{\bf45.3}&{\bf56.8}\\

\hline

\end{tabular}
\end{center}
\label{r_h36mshort2}
\vspace{-2.5em}
\end{table*}

\textbf{Human3.6M} \cite{24} is the most widely used benchmark for motion prediction. It involves 15 actions performed by professionals, and each human pose involves a 32-joint skeleton. Following \cite{08,20}, we compute the joint's 3D coordinates by applying forward kinematics and down-sample the motion sequence to 25 frames per second. After removing the global rotation, translation and constant 3D coordinates of each human pose, there remains 22 joints. We test our method on subject 5(S5). Considering \cite{32} and \cite{31} show the setting of 8 random sub-sequences may lead to high variance, we test our model with two different division methods in previous works. One is 8 random sub-sequences per action on subject 5 (S5) and another is 256 sub-sequences per action.

\textbf{3DPW} \cite{25} The 3D Pose in the Wild dataset(3DPW) \cite{25} consists of challenging indoor and outdoor actions. The dataset consists of various activities such as shopping, doing sports, and hugging, including 60 sequences and more than 51k frames. For a fair comparison, we evaluate the whole test sets.

\textbf{CMU-Mocap} \cite{-10} The CMU mocap dataset mainly includes five categories, naming ``human interaction'', ``interaction with environment'', ``locomotion'', ``physical activities \& sports'' and ``situations \& scenarios''. Be consistent with \cite{08,20}, we select 8 detailed actions: ``basketball'', ``basketball signal'', ``directing traffic'', ``jumping'', ``running'', ``soccer'', ``walking'' and ``washing window''. We evaluate our model with the same approach as we do for H3.6M.

\textbf{Network Setting}. We take three timescales: 1, 3, and 5 frames around the target frame in MTDE. The size of the high-level dimension $D$ and $T$ is 32 and 64. To get enough receptive field, we adapt nine stacked CJRE modules with the lateral connection.

\textbf{Training}. All training is conducted on the pytorch platform with one 2080Ti GPU. We use Adam \cite{26} optimizer with an initial learning rate of 0.0005. We use a weight decay of 0.96 and set the learning rate as 0.0001 at the epoch 4. The batch size is set to 16.

\begin{table*}[!t] 
\caption{Long-term prediction of 256 sub-sequences per action on H$3.6$M. Where ``ms'' denotes ``milliseconds''.}
\scriptsize
\begin{center}
\begin{tabular}{c|cccc|cccc|cccc|cccc}
\hline
motion & \multicolumn{4}{c}{Walking} & \multicolumn{4}{c}{Eating}& \multicolumn{4}{c}{Smoking} & \multicolumn{4}{c}{Discussion}\\
\hline
time(ms)&560&720&880&1000&560&720&880&1000&560&720&880&1000&560&720&880&1000 \\
\hline
ResSup    \cite{17} &71.6 &72.5 &76.0 &79.1         & 74.9 &85.9 &93.8 &98.0            & 78.1 & 88.6 & 96.6 &102.1                   & 109.5 & 122.0 &128.6 &131.8 \\
ConvS2S   \cite{19} &72.2 &77.2 &80.9 &82.3         & 61.3 &72.8 &81.8 &87.1            & 60.0 & 69.6 & 77.2 &81.7                      & 98.1  & 112.9 &123.0 &129.3\\

LTD     \cite{08} &51.8 &56.2 &58.9 &60.9         & 50.0 &61.1 &69.6 &74.1          & 51.3  & 60.8 & 68.7 &73.6                   & 87.6 & 103.2 &113.1 &118.6 \\
HRI     \cite{31} &{\bf47.4} &{\bf52.1} &{\bf55.5} &{\bf58.1}         & 50.0 &61.4 &70.6 &75.7            & {\bf47.6}  & {\bf56.6} & {\bf64.4} &{\bf69.5}                 & 86.6 & 102.2 &113.2 &119.8 \\

\hline
 Ours&47.9&52.5&56.2&59.6   &{\bf47.6}&{\bf59.2}&{\bf68.1}&{\bf73.2}    &48.4&57.7 &64.9&69.6   &{\bf85.3} &{\bf100.5}&{\bf110.0}&{\bf115.3}\\

\hline
\hline
motion & \multicolumn{4}{c}{Direction} & \multicolumn{4}{c}{Greeting}& \multicolumn{4}{c}{Phoning} & \multicolumn{4}{c}{Posing}\\
\hline
time(ms)&80&160&320&400&80&160&320&400&80&160&320&400&80&160&320&400 \\
\hline
ResSup    \cite{17} &101.1 &114.5 &124.5 &129.1         & 126.1 &138.8 &150.3 &153.9            & 94.0  & 107.7 & 119.1 &126.4          & 140.3 & 159.8 &173.2 &183.2 \\
ConvS2S   \cite{19} &86.6  &99.8  &109.9 &115.8         & 116.9 &130.7 &142.7 &147.3            & 77.1  & 92.1  & 105.5 &114.0          & 122.5 & 148.8 &171.8 &187.4\\

LTD       \cite{08} &76.1  &91.0 &102.8 &108.8         & 104.3 &120.9 &134.6 &140.2             & 68.7  & 84.0  & 97.2  &105.1          & 109.9 & 136.8 &158.3 &171.7 \\
HRI     \cite{31} &73.9  &88.2 &100.1 &106.5         & 101.9 &118.4 &132.7 &138.8           & 67.4  & 82.9  & 96.5  &{\bf105.0}                 & 107.6 & 136.8 &161.4 &178.2 \\

\hline
 Ours&{\bf72.8}&{\bf87.3}&{\bf98.4}&{\bf104.6}    &{\bf99.4}&{\bf115.8}&{\bf128.7}&{\bf134.6}    &{\bf66.2}&{\bf82.2} &{\bf96.2}&105.1    &{\bf105.3} &{\bf113.4}&{\bf115.9}&{\bf170.4}\\

\hline
\hline
motion & \multicolumn{4}{c}{Purchasing} & \multicolumn{4}{c}{Sitting}& \multicolumn{4}{c}{Sitting down} & \multicolumn{4}{c}{Taking photo}\\
\hline
time(ms)&560&720&880&1000&560&720&880&1000&560&720&880&1000&560&720&880&1000 \\
\hline
ResSup    \cite{17} &112.1 &137.2 &148.0 &154.0             & 113.7 &130.5 &144.4 &152.6            & 138.8 & 159.0 & 176.1 &187.4            & 110.6 & 128.9 &143.7 &153.9 \\
ConvS2S   \cite{19} &111.3 &129.1 &143.1 &151.5             & 82.4  &98.8  &112.4 &120.7            & 106.5 & 125.1 & 139.8 &150.3            & 84.4  & 102.4 &117.7 &128.1\\

LTD       \cite{08} &99.4  &114.9 &127.9 &135.9             & 78.5  &95.7  &110.0 &118.8          & 99.5  & 118.5 & 133.6 &144.1            & 76.8  & 95.3 &110.3 &120.2 \\
HRI     \cite{31} &{\bf95.6}  &{\bf110.9} &125.0 &134.2       & 76.4  &93.1  &107.0 &115.9            & 97.0  & 116.1 & 132.1 &143.6                  & {\bf72.1}  & 90.4 &105.5 &115.9 \\

\hline
 Ours&96.2&111.5&{\bf124.3}&{\bf131.6}                &{\bf75.1}&{\bf92.0}&{\bf105.6}&{\bf113.8}    &{\bf94.9}&{\bf114.6} &{\bf131.1}&{\bf142.5}    &72.3 &{\bf90.3}&{\bf105.0}&{\bf113.8}\\
\hline
\hline
motion & \multicolumn{4}{c}{Waiting} & \multicolumn{4}{c}{Walking dog}& \multicolumn{4}{c}{Walking Together} & \multicolumn{4}{c}{Average}\\
\cline{1-17}
time(ms)&560&720&880&1000&560&720&880&1000&560&720&880&1000&560&720&880&1000 \\
\hline
ResSup    \cite{17} &105.4 &117.3 &128.1 &135.4         & 128.7 &141.1 &155.3 &164.5            & 80.2 & 87.3 & 92.8 &98.2          & 106.3 & 119.4 &130.0 &136.6 \\
ConvS2S   \cite{19} &87.3  &100.3 &110.7 &117.7         & 122.4 &133.8 &151.1 &162.4            & 72.0 & 77.7 & 82.9 &87.4            & 90.7  & 104.7 &116.7 &124.2\\

LTD       \cite{08} &75.1  &88.7  &99.5  &106.9         & 105.8 &118.7 &132.8 &142.2          & 58.0 & 63.6 & 67.0 &69.6          & 79.5  & 94.0  &105.6 &112.7 \\
HRI     \cite{31} &74.5  &89.0  &100.3 &108.2         & 108.2 &120.6 &135.9 &146.9            & {\bf52.7} &{\bf57.8} & 62.0 &64.9                  & 77.3  & 91.8  &104.1 &112.1 \\

\hline
 Ours&{\bf69.7}&{\bf83.4}&{\bf95.0}&{\bf102.4}    &{\bf102.9}&{\bf115.2}&{\bf131.8}&{\bf141.6}    &53.1&58.1 &{\bf61.3}&{\bf63.8}    &{\bf75.8} &{\bf90.3}&{\bf102.2}&{\bf109.5}\\

\hline

\end{tabular}

\end{center}
\label{r_h36mlong2}
\vspace{-2.5em}
\end{table*}

\begin{table*}[htb] 
\caption{Short-term and long-term prediction on CMU-mocap.}
\scriptsize
\begin{center}
\begin{tabular}{c|ccccc|ccccc|ccccc}
\hline
motion& \multicolumn{5}{c}{basketball} & \multicolumn{5}{c}{baskeball Signal}& \multicolumn{5}{c}{Directing Traffic}\\
\hline
time (ms) & 80 &160 & 320 &400 &1000& 80 &160 & 320 &400 &1000& 80 &160 & 320 &400 &1000\\
\hline
LTD \cite{08}&14.0&25.4 & 49.6 &61.4&{\bf106.1} &3.5 &  6.1 & 11.7 &15.2 &  53.9& 7.4 & 15.1 &31.7 &  42.2 &152.4\\

TrajCNN \cite{20}&11.1&19.7 &43.9&  56.8& 114.1&  {\bf1.8}  &3.5  &{\bf9.1} &13.0 &{\bf49.6}  &{\bf5.5} & {\bf10.9} &{\bf23.7} &{\bf31.3}&  {\bf105.9}\\
\hline
Ours&{\bf11.1}&{\bf19.5}  &{\bf42.8}& {\bf55.7}&  113.1&  1.9 &{\bf3.5} &9.3  &{\bf13.0}  &57.5 &5.8 &  11.7  &25.6&33.4& 139.0\\
\hline
motion& \multicolumn{5}{c}{Jumping} & \multicolumn{5}{c}{Running}& \multicolumn{5}{c}{Soccer}\\
\hline
time (ms) & 80 &160 & 320 &400 &1000& 80 &160 & 320 &400 &1000& 80 &160 & 320 &400 &1000\\
\hline
LTD \cite{08}&16.9&   34.4 &  76.3  &96.8   &164.6  &25.5   &36.7&  39.3 &  39.9  &58.2 &11.3   &21.5   &44.2&  55.8 &  117.5\\
TrajCNN \cite{20}&12.2  &28.8&  {\bf72.1}&  94.6& 166.0 &17.1&  24.4  &28.4&  32.8& 49.2  &{\bf8.1} &{\bf17.6}& 40.9  &51.3 &126.5\\
\hline
Ours&{\bf11.4}  &{\bf28.0}& 72.7& {\bf94.1}&  {\bf155.3}  &{\bf16.4}& {\bf20.1} &{\bf22.9}& {\bf27.6}&  {\bf41.9} &8.6  &18.3&  {\bf39.1} &{\bf48.4}  &{\bf103.6}\\
\hline

\end{tabular}
\begin{tabular}{c|ccccc|ccccc|ccccc}
\hline
motion& \multicolumn{5}{c}{Walking} & \multicolumn{5}{c}{Wash Window}& \multicolumn{5}{c}{Average}\\
 \cline{1-16}
time (ms) & 80 &160 & 320 &400 &1000& 80 &160 & 320 &400 &1000& 80 &160 & 320 &400 &1000\\
\hline
LTD \cite{08}&7.7 & 11.8 &  19.4 &  23.1 &  40.2  &5.9 &  11.9 &  30.3 &  40.0 &  79.3  &11.5 & 20.4 &  37.8 &  46.8 &  96.5\\
TrajCNN \cite{20}&6.5 &10.3&  19.4& 23.7& 41.6  &{\bf4.5}&{\bf  9.7}  &29.9&  41.5& 89.9  &8.3  &15.6&  33.4  &43.1 &92.8\\
\hline
Ours&{\bf5.9} &{\bf9.0}&  {\bf17.4}&  {\bf21.1}&  {\bf38.8}&  4.6 &10.0 &{\bf28.6}  &{\bf39.0}& {\bf73.1} &{\bf8.2}&  {\bf15.1} &{\bf32.3}& {\bf41.5} &{\bf90.3}\\
\hline
\end{tabular}
\end{center}
\label{results_cmu}
\vspace{-2.5em}
\end{table*}

\subsection{Comparison with baselines}
Here we show the prediction performance for both short-term and long-term motion prediction on Human3.6M (H3.6M), CMU-Mocap and 3DPW. We quantitatively evaluate various methods by the mean per joint position error (MPJPE) between the generated motions and ground truths. To be consistent with the literature \cite{20,08}, we report our results for short-term ($<$ 500ms) and long-term ($>$ 500ms) predictions. For all datasets, we are given 10 frames (400 milliseconds) to predict the future 10 frames (400 milliseconds) for short-term prediction and to predict the future 25 frames (1 second) for long-term prediction.
\subsubsection{Results on H3.6M}

TABLE \ref{r_h36mshort} provides the short-term prediction of 8 sub-sequences per action on H3.6M for the 15 activities and the average results. Note that our method outperforms all the baselines on average and almost all motions, which indicates our proposed framework's effectiveness. These possible reasons are twofold: (1) Our model extracts better motion features with our proposed joint relation modeling. These features contain both global coordination of all joints and local interactions between joint pairs and thus could offer more reliable guidance for effective prediction. (2) Our model extracts enriched motion dynamics in MTDE, which provide more motion cues for later motion prediction. Thus, our model generally outperforms the listed baselines in almost all actions. Specifically, for those motions that need the upper body and lower body to cooperate, e.g., ``Walking dog'', ``Phoning'' and ``Sitting down'', our method outperforms most, which reflects the efficacy of our proposed global coordination of all joints. These motions all need the whole body to participate in, and thus global coordination could offer more reliable guidance. Besides, the result on 320ms and 400ms increase most, which shows our method is good at capturing temporal continuity for long-term prediction compared with other methods.

In TABLE \ref{r_h36mlong}, we compare our model with other baselines for long-term prediction of 8 sub-sequences per action on H3.6M. With the uncertainly of motion increasing, our method still obtains competitive performances on almost all motions. In TABLE \ref{r_h36mshort2} and TABLE \ref{r_h36mlong2}, we also show the short-term and long-term prediction of 256 sub-sequences per action on H3.6M for the 15 activities and the average results. We can find that with the number of test data increasing, our method still outperforms all baselines in almost all motions. It proves that our method has great generalization in different situations, whether the samples are big or not. The above observations all demonstrate the advantages of our proposed enriched dynamics and comprehensive joint relation modeling.

\begin{table}[htb]
\caption{Short and long-term prediction on 3DPW. }
\scriptsize
\begin{center}
\centering
\begin{tabular}{c|ccccc}
\hline
{time (ms)} & 200 &400 & 600 &800 & 1000\\
 \hline
{LTD} \cite{08} & 36.0 &69.0 &91.0 &107.6 &118.6\\
\hline
{Ours} &{\bf 34.7} &{\bf 66.7} &{\bf 85.6}&{\bf 98.0} &{\bf 108.4}\\
 \hline
\end{tabular}
\end{center}
\label{r3dpw}

\end{table}

\subsubsection{Results on CMU Mocap and 3DPW}
TABLE \ref{results_cmu} reports the MPJPE for short-term and long-term prediction on CMU-Mocap and TABLE \ref{r3dpw} reports the results on 3DPW. Essentially, the conclusions remain unchanged: our method consistently outperforms the baselines for both short-term and long-term prediction.

\begin{figure*}[h] 
\begin{center}
\includegraphics[width=0.8 \textwidth]{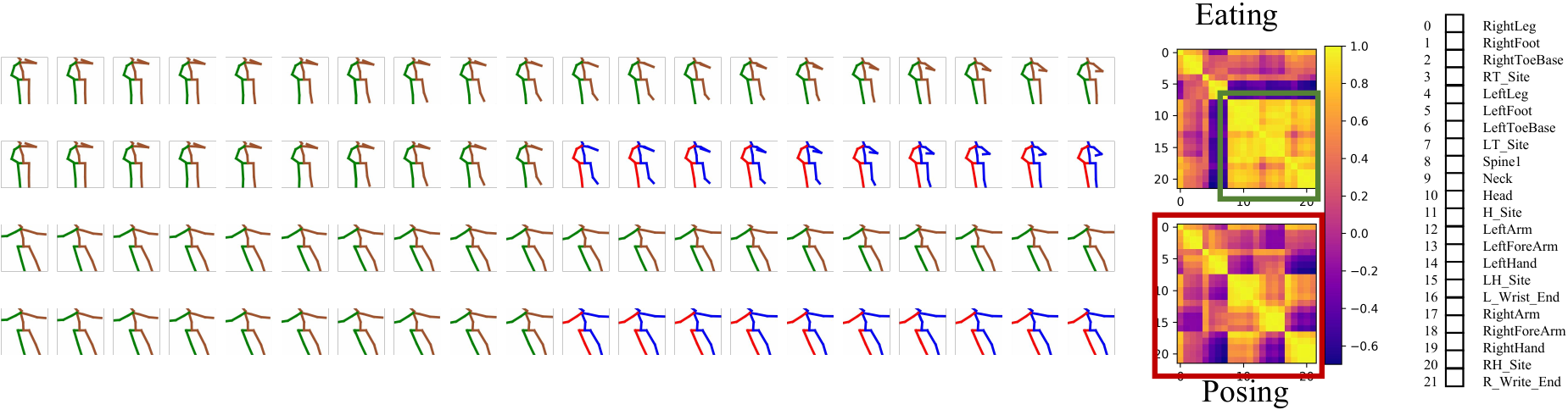}
\end{center}
\caption{The middle panel is the visualizations of one feature map of $C_{emb}$, of which the size is $[22\times 22]$. The right panel is the corresponding prediction results. The left panel is the annotation of the feature map. From top to bottom, we show the ground truth and our approach of two different motions.}
\label{f_relation_vector}
\end{figure*}

\subsection{Ablation study}
In this section, we conduct several ablation experiments to testify the effectiveness of different components in our proposed framework. The models are trained on the H3.6M training set and evaluated on the test dataset. The comparison results are shown in TABLE \ref{ab_1}, \ref{ab_2}, \ref{ab_3}, \ref{ab_5}, \ref{ab_4}.

\subsubsection{Multi-timescale dynamics}

The multi-timescale dynamics could offer enriched motion cues than position or velocity information provided by the raw input motion sequences. And we design the MTDE module to encode this information in our work. As is shown in Table \ref{ab_1}, we design two extra experiments. ``TS'' means that we only adopt two-stream input and ``TS+FE'' means that we intorduce bare 1*3 convolution without extra designs. Our final scheme is ``TS+FE+MT'' which adds the multiple temporal convolutions compared with ``TS+FE''. We can see ``TS+FE'' is superior to ``TS'', which indicates that more motion cues are useful. Besides, after using ``TS+FE+MT'', we could get a fairly good promotion by adopting the MTDE module in all time horizons. Significantly, the results of 320ms and 400ms increase a lot, showing that enriched dynamics could offer more meaningful guidance for middle or long-term prediction.

\begin{table}[h]
	
	\scriptsize
	\begin{center}
		\centering
		\caption{Results of ablation experiments on MTDE}
		\label{ab_1}
		\begin{tabular}{c|cccc}
			\hline
			MTDE&80&160&320&400\\
			\hline
			TS & 10.2 &22.9 &49.5 &	60.6 \\
			TS+FE & 9.8 &22.6 &48.0 &58.4 \\
			TS+FE+MT &{\bf9.6} &{\bf22.0} &{\bf46.3} &{\bf57.0}\\
			\hline
		\end{tabular}
	\end{center}
	
\end{table}

\subsubsection{Global coordination of all joints}

The global coordination of all joints could reflect the entirety and coordination of the predicted motion and thus is one vital joint relation. In our work, we propose the GCE module to model this global coordination. We here illustrate the effectiveness of several special designs in this module. In the Feature Normalization  (FNU), relative joint motion representation is a vital design used to remove the effect of global motion trends because the global coordination essentially reflects the mutual constraints of all joints. Here, we denote “$RJ$” to represent the usage of relative joint motion representations. In the Multi-head Self-attention Unit (MSU), we generate multiple relation graphs to extract richer motion coordination. Here we use “$MR$” to represent the usage of multiple relation graphs. Besides, In the MSU, we also choose cosine similarity to calculate the joint relations in our paper and illustrate the advantage of this choice. To prove the effectiveness, we design an experiment with the softmax function as a comparison. Here “$Sim_c$” and “$Sim_s$” represent the usage of cosine similarity or softmax.

\begin{table}[h]
\caption{Results of ablation experiments on GCE}
\scriptsize
\begin{center}
\centering
\begin{tabular}{cccc|cccc}
\hline
$RJP$&$MR$&$Sim_c$&$Sim_s$&80&160&320&400\\
\hline

\Checkmark&\Checkmark&&\Checkmark& 10.2 &23.4 &49.5&60.6\\
&\Checkmark&\Checkmark&& 9.7 &22.3 &47.4 &58.4 \\
\Checkmark&&\Checkmark&& 9.8 &22.1 &46.7 &57.7 \\
\Checkmark&\Checkmark&\Checkmark&&{\bf9.6} &{\bf22.0} &{\bf46.3} &{\bf57.0}\\
\hline
\end{tabular}
\end{center}
\label{ab_2}
\end{table}

(1) By adopting the relative joint representation, the final performance outperforms 0.1, 0.3, 1.1, 1.4 for 80ms, 160ms, 320ms, 400ms, respectively. This proves that the relative joint representation without the global motion trends is more suitable for modeling joints' global coordination. 

(2) We can find the results are better with the usage of multiple relation graphs. It demonstrates that combining different relations is beneficial to get richer and more diverse coordination.

(3) The cosine similarity is better compared with the softmax function used in computing self-attention. It arises from two aspects. First, it avoids violent differences in the softmax function because cosine similarity limits the value in $[-1,1]$. Second, it has the angle information to represent both orientation and intensity of correlation, while softmax only represents the intensity of correlation.

\begin{table}[h]
\caption{Results of ablation experiments on GCE, LIE and AFFM}
\scriptsize
\begin{center}
\centering

\begin{tabular}{ccc|cccc}
\hline
$GCE$&$LIE$&AFFM&80&160&320&400\\
\hline
\Checkmark& &\Checkmark&  9.7 &22.6 &48.3 &58.9\\

&\Checkmark &\Checkmark& 10.1 &23.1 &49.2 &60.0 \\
\Checkmark&\Checkmark & & 9.6 &22.4 &46.8 &57.4 \\

\Checkmark&\Checkmark &\Checkmark&{\bf9.6} &{\bf22.0} &{\bf46.3} &{\bf57.0}\\

\hline
\end{tabular}
\end{center}
\label{ab_3}

\end{table} 

\begin{figure*}[htb] 
\centering 
\subfigure[Sitting 0°]{\label{fig:subfig:a2}
\includegraphics[width=0.8\linewidth]{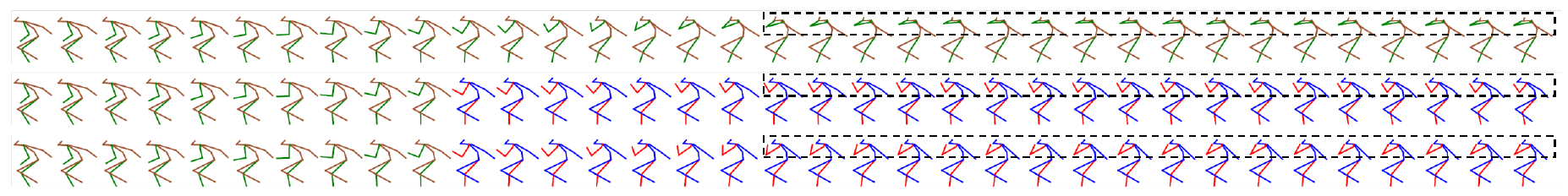}}
\subfigure[Sitting 30°]{\label{fig:subfig:b2}
\includegraphics[width=0.8\linewidth]{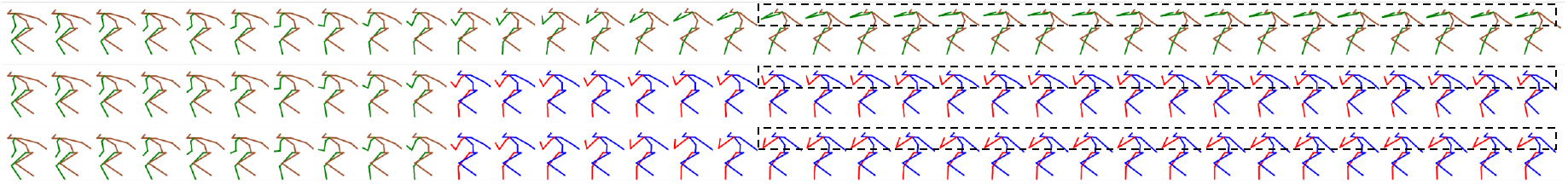}}
\subfigure[Sitting 90°]{\label{fig:subfig:b2}
\includegraphics[width=0.8\linewidth]{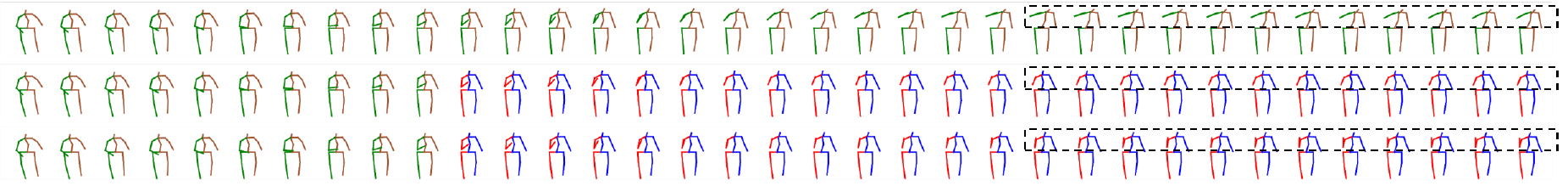}}
\subfigure[Sitting 120°]{\label{fig:subfig:b2}
\includegraphics[width=0.8\linewidth]{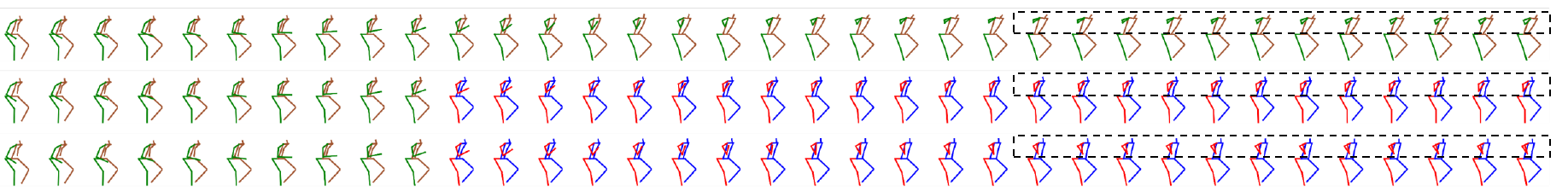}}
\caption{Qualitative comparison of multi-view long-term predictions on H3.6M. From top to bottom, we show the ground truth, the results of LTD \cite{08}, and our approach. The highlight illustrates the difference between three sequences.}
\label{f2}
\end{figure*}

\begin{figure}[htb] 
\centering 
\subfigure[Walking dog 0°]{\label{fig:subfig:a1}
\includegraphics[width=1\linewidth]{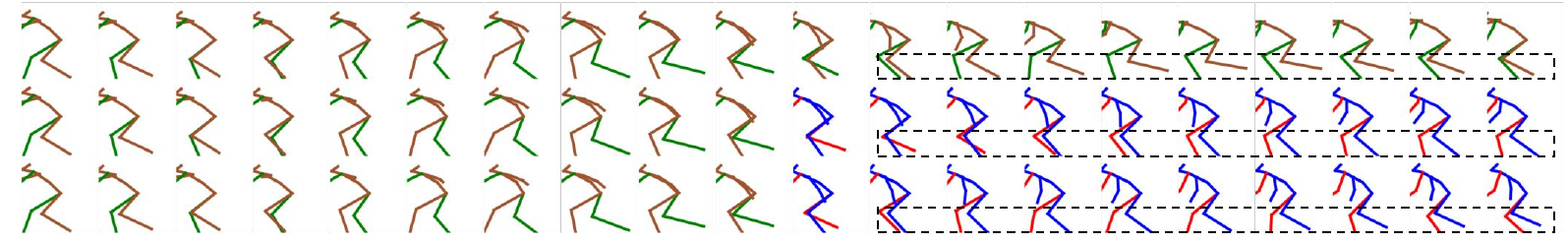}}
\subfigure[Walking dog 30°]{\label{fig:subfig:b1}
\includegraphics[width=1\linewidth]{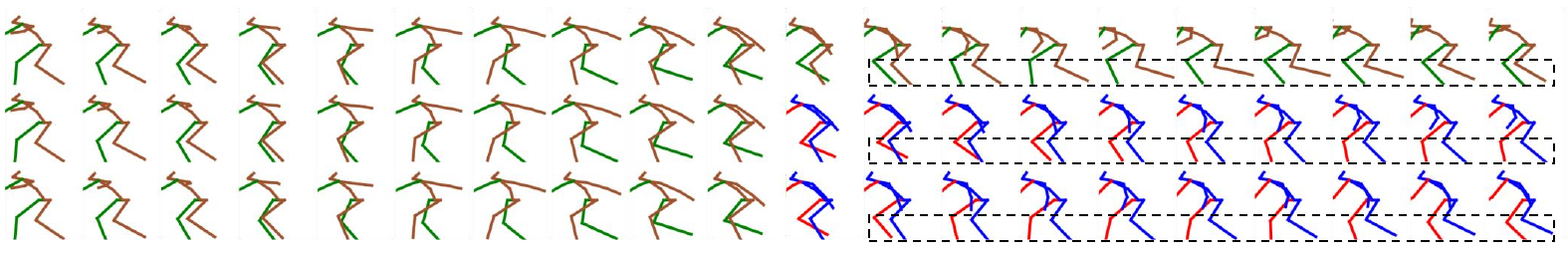}}
\subfigure[Walking dog 90°]{\label{fig:subfig:c1}
\includegraphics[width=1\linewidth]{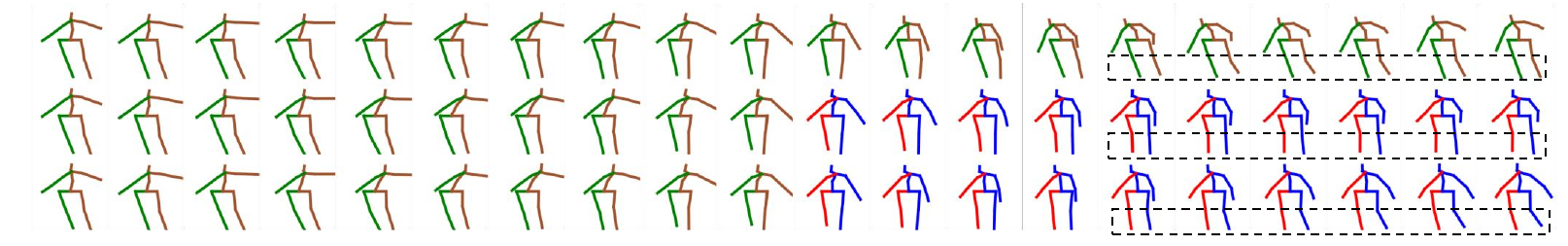}}
\subfigure[Walking dog 120°]{\label{fig:subfig:d1}
\includegraphics[width=1\linewidth]{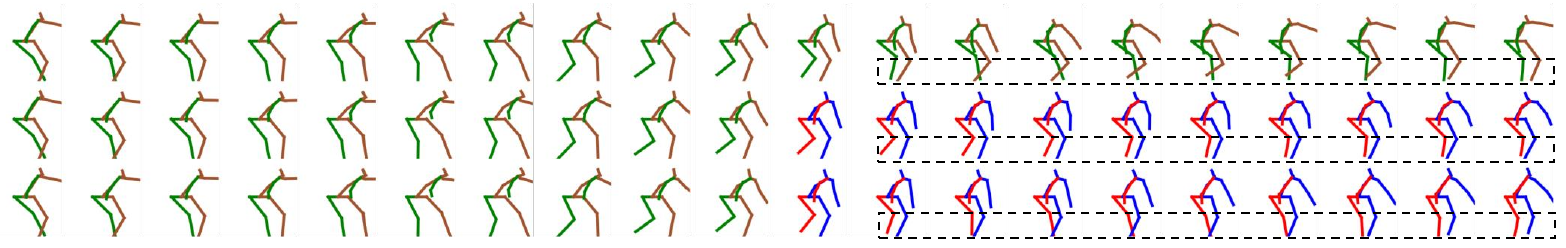}}
\caption{Qualitative comparison of multi-view short-term predictions on H3.6M. From top to bottom, we show the ground truth, the results of LTD \cite{08}, and our approach. The highlight illustrates the difference between three sequences. The results evidence that our approach generates high-quality predictions.}
\label{f1}
\end{figure}

\begin{figure*}[htb] 
\centering 
\subfigure[Running 0°]{\label{fig:subfig:a3}
\includegraphics[width=0.7\linewidth]{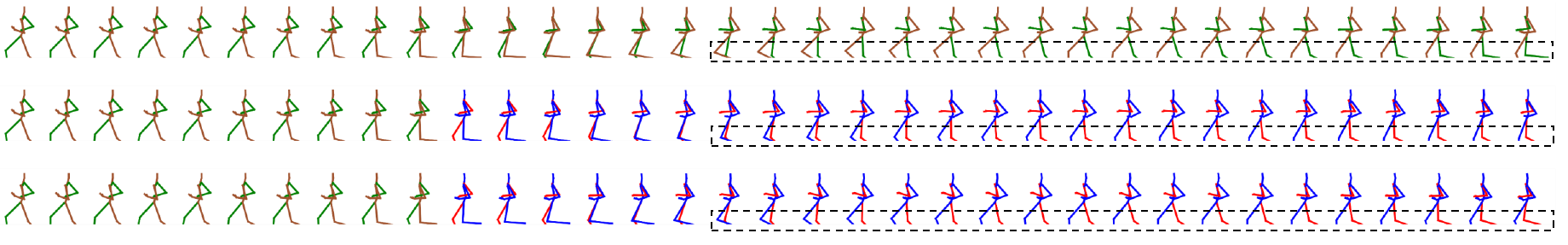}}
\subfigure[Running 30°]{\label{fig:subfig:b3}
\includegraphics[width=0.7\linewidth]{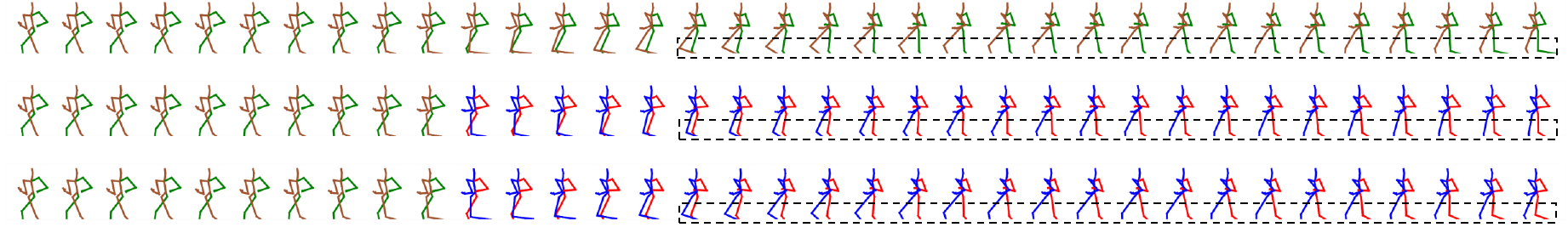}}
\subfigure[Running 90°]{\label{fig:subfig:c3}
\includegraphics[width=0.7\linewidth]{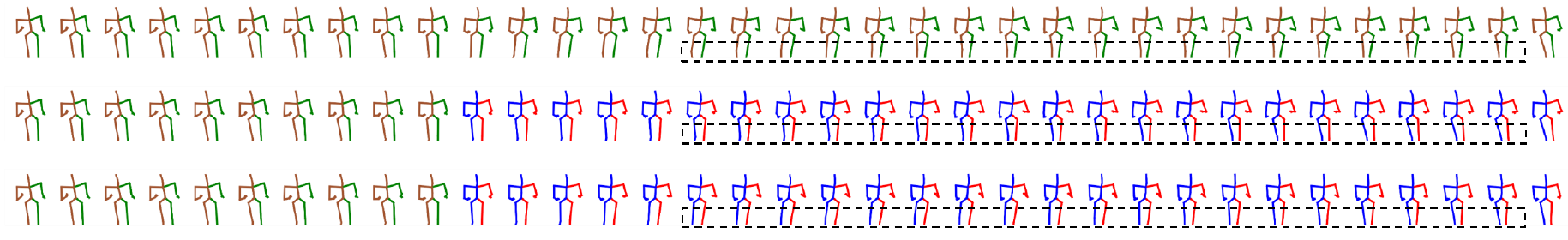}}
\subfigure[Running 120°]{\label{fig:subfig:c3}
\includegraphics[width=0.7\linewidth]{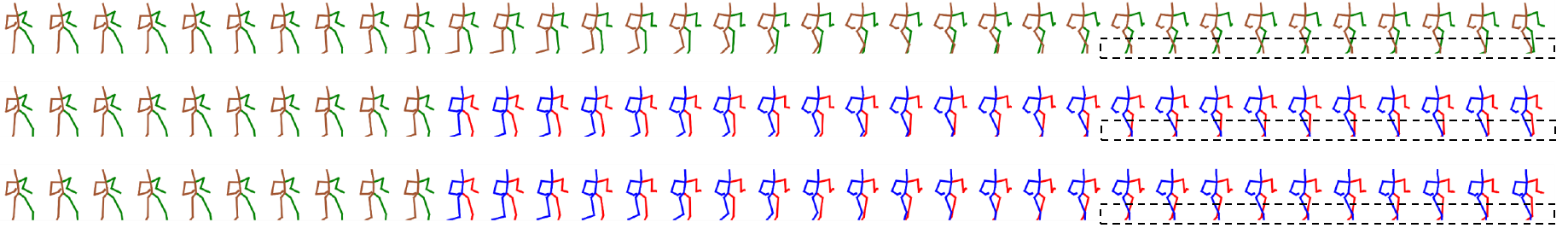}}
\caption{Qualitative comparison of multi-view short-term and long-term predictions on CMU-Mocap. From top to bottom, we show the ground truth, the results of LTD \cite{08}, and our approach. The highlight illustrates the difference between three sequences.}
\label{f3}
\end{figure*}

\subsubsection{Importance of different joint relations}
The global coordination of all joints and local interactions between joint pairs are complementary and crucial joint relations. In TABLE \ref{ab_3}, we could easily find that these two joint relations can promote each other, and adopting both of them can achieve better improvement. Notably, the method with a single GCE outperforms the one with a single LIE. This observation demonstrates that global coordination could extract more cues and offer more effective guidance than local interactions in prediction. 

Furthermore, we also conduct the experiments in TABLE \ref{ab_5} to verify the performance of different combination method of GCE and LIE. We denote the “Parallel” as putting the GCE and LIE in the parallel path and “Serial” means that local interaction is conditioned on global coordination. We can see that if we set the LIE behind the GCE, we won’t get equivalent performance compared with putting the GCE and LIE in parallel. The main reason is the output features of GCE are not suitable to act as the input of LIE. More concretely, the output features of GCE represent the joint features relative to the CA. Thus, the adjacent pixels in the new relative feature maps lost the adjacent relations existing in the raw skeletal structure, which is the prerequisite of the LIE module.

\begin{table}[h]
	\scriptsize
	\begin{center}
		\centering
		\caption{Results of on the combination method of GCE and LIE}
		\label{ab_5}
		\begin{tabular}{c|cccc}
			\hline
			combination method&80&160&320&400\\
			\hline
			Serial & 10.5 &23.5 &49.6 &	60.4 \\
			Parallel(Ours) &{\bf9.6} &{\bf22.0} &{\bf46.3} &{\bf57.0}\\
			\hline
		\end{tabular}
	\end{center}
	
\end{table}

Besides, we consider that different motions may have different preferences for those above two joint relations and thus design the AFFM module. As is shown in TABLE \ref{ab_3}, AAFM improves the results by 0.4 on average. It reflects that fusing the motion features achieved from different joint relations enhances the whole performance. Notably, the limited improvement of AFFM reflects that the joint coordination modeling is the key design to improve the final prediction performance.

\begin{table}[h]
	
	\scriptsize
	\begin{center}
		\centering
		\caption{Relative weight of different features of different motions in AFFM}
		\label{ab_4}
		\begin{tabular}{c|ccc}
			\hline
			$Motions$&$w_{distant}$&$w_{adjacent}$&$w_{ca}$\\
			\hline
			walking &0.26 &0.40  &0.34\\
			\hline
			eating &0.32& 0.36& 0.33 \\
			\hline
			smoking &0.35& 0.33& 0.31 \\
			\hline
			discussion &0.31& 0.36& 0.33\\
			\hline
			directions &0.31&0.36&0.33 \\
			\hline
			greeting &0.30& 0.36& 0.34\\
			\hline
			phoning &0.31& 0.37&   0.32\\
			\hline
			posing &0.35& 0.33& 0.33\\
			\hline
			purchases &0.31&  0.36& 0.33 \\
			\hline
			sitting &0.34&  0.34&  0.32\\
			\hline
			sittingdown &0.35& 0.33& 0.32\\
			\hline
			takingphoto &0.33& 0.34&0.33 \\
			\hline
			waiting &0.29& 0.37&0.34 \\
			\hline
			walkingdog &0.28& 0.38& 0.34\\
			\hline
			walkingtogether &0.27& 0.39&0.34\\
			\hline
		\end{tabular}
	\end{center}

\end{table} 

To demonstrate how LIE and GCE are weighted in AFFM module, we here offer the relative weight of different features in different motions. As is shown in Table\ref{ab_4}, ${w_{distant}}$, ${w_{adjacent}}$, ${w_{ca}}$ represent the relative weight of ${F_{distant}}$, ${F_{adjacent}}$, ${F_{ca}}$ respectively. From the results, we can find the ratio of global motion coordination ${w_{ca}}$ varies a little, which means that the global motion coordination is of almost equal importance in all kinds of motions. Besides, we can also find ${w_{distant}}$ and ${w_{adjacent}}$ are usually contrary. For those motions with little movement like ``smoking'', ``sitting down'' and ``posing'', the importance of $F_{distant}$ is higher. For those motions with more movement like ``walking’, ``walking dog'' and ``walking together'', the importance of $F_{adjacent}$ is higher. It demonstrates that dynamic motions may focus on the movement of local bodies while static motions pay more attention to the overall structure of the human body.

\subsection{Qualitative analysis}

\subsubsection{Visulization of different motions on H3.6M and CMU Mocap}

As is shown in Fig. \ref{f2}, Fig. \ref{f1}, and Fig. \ref{f3}, we list more qualitative visualizations of different datasets. We can find that our method outperforms LTD\cite{08} in both short-term and long-term prediction. In detail, compared with LTD\cite{08}, it is easier for our model to capture motion tendency and make more precise predictions. For example, in Fig. \ref{f1}, we can see that the movement of the left leg are well learned from four perspectives in our prediction for the motion ``Walking dog'' while not in LTD\cite{08}. And in the motion  ``Sitting'', our model is more precise in predicting upper limb movements from multiple viewpoints. It demonstrates that our methods can better capture the mutual constraints of different body parts, and thus the resulting predicted motion appears more natural and realistic. Besides, for those motions  like ``Running'', our results usually outperform other methods, which shows the enriched dynamics can offer more useful motion cues for motion prediction.

\subsubsection{Visulization of global coordination}

In Fig. \ref{f_relation_vector}, we show the one feature map of $C_{emb}$, which reflects the global coordination of all joints. The value increases with the color become brighter. For the motion ``Eating'', there exist high correlations on the upper body because the joints on the upper body need to coordinate to finish the motion. While for motion ``Posing'' there exist fewer correlations between different body parts because this motion is more static than ``Eating''. Thus, the coordination only occurs in the local body parts. This observation demonstrates that our proposed GCE extract reliable global coordination for effective prediction.

\section{Conclusion}

In this paper, we focus on more realistic human motion prediction with attention to motion coordination. To this end, we propose the CJRE to explore richer joint relation modeling, mainly including GCE and LIE. The former presents the global coordination of all joints and the latter encodes local interactions between joint pairs. Besides, we also extract enriched motion dynamics of raw skeleton data through the MTDE to exploit more motion cues for effective prediction. Experimental results on three benchmark datasets suggest that our proposed framework is able to improve the coordination of predicted motions with lower errors to generate more realistic actions.


\section*{Acknowledgment}

This work was supported partly by the National Natural Science Foundation of China (Grant No. 62173045, 61673192), and partly by the Fundamental Research Funds for the Central Universities(Grant No. 2020XD-A04-2).

\ifCLASSOPTIONcaptionsoff
  \newpage
\fi



%
\bibliographystyle{IEEEtran}
\bibliography{egbib}

%

\begin{IEEEbiography}[{\includegraphics[width=1in,height=1.25in,clip,keepaspectratio]{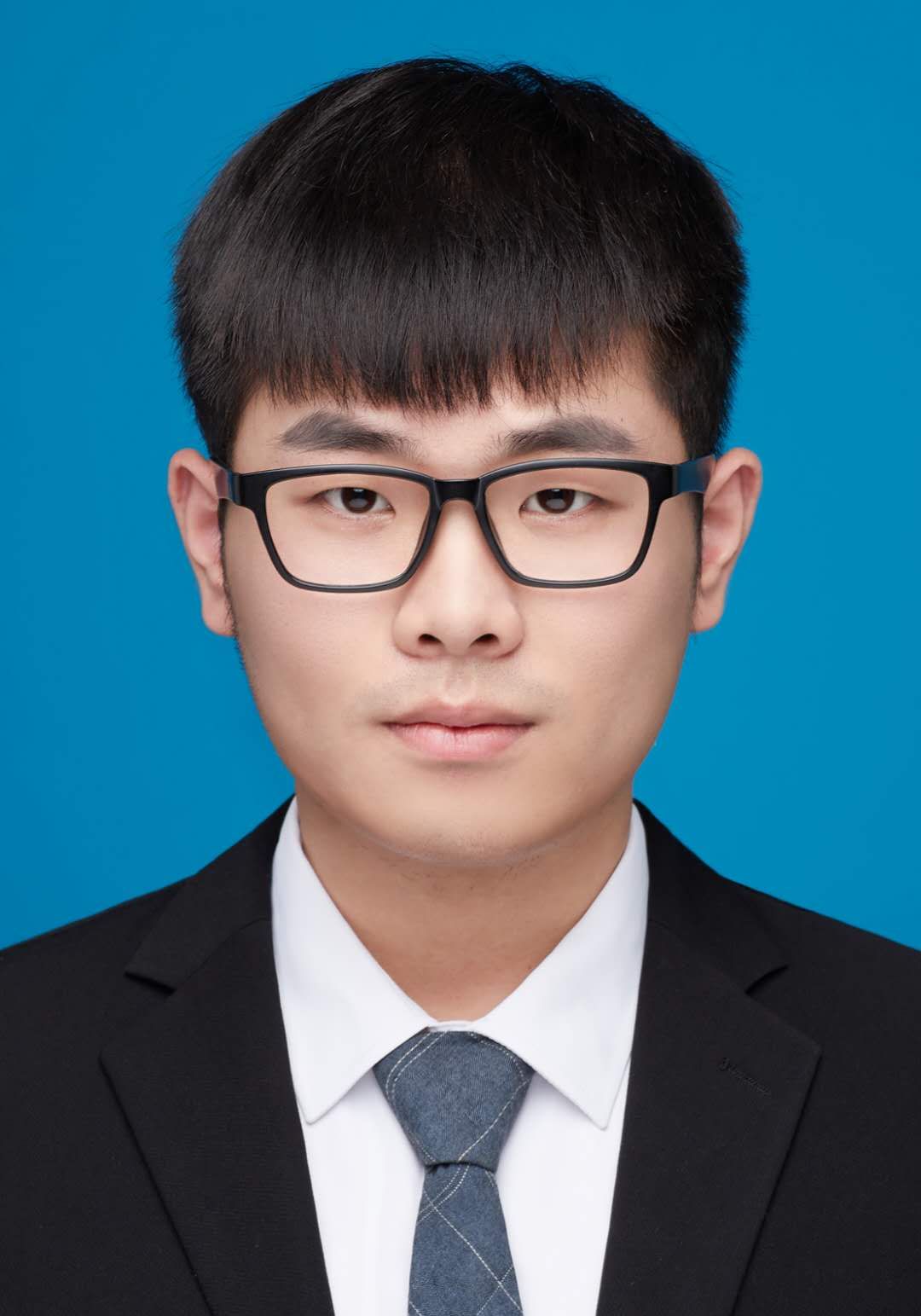}}] {Pengxiang Ding} received the B.S. degree from the
Beijing University of Post and Telecommunications,
Beijing, China, in 2019, where he is currently pursuing the M.S. degree with the School of Artificial
Intelligence. His research interests include motion
prediction and action recognition in computer vision.
\end{IEEEbiography}


\begin{IEEEbiography}[{\includegraphics[width=1in,height=1.25in,clip,keepaspectratio]{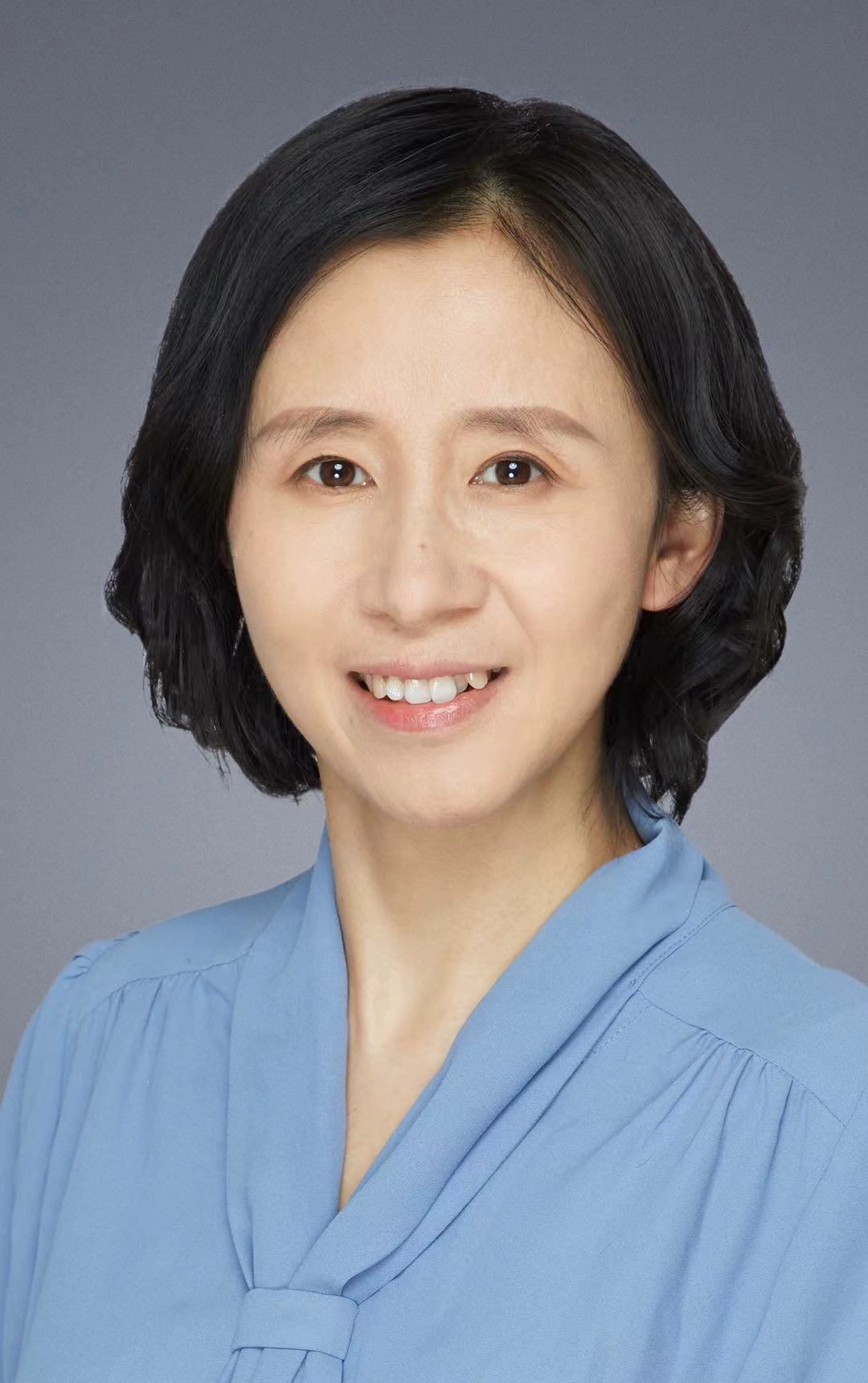}}]{Jianqin Yin}(Member, IEEE) received the Ph.D. degree from Shandong University, Jinan, China, in 2013. She currently is a Professor with the School of Artificial Intelligence, Beijing University of Posts and Telecommunications, Beijing, China. Her research interests include service robot, pattern recognition, machine learning, and image processing.
\end{IEEEbiography}






\end{document}